\documentclass[conference,compsoc]{IEEEtran}

\ifCLASSOPTIONcompsoc
  \usepackage[nocompress]{cite}
\else
  \usepackage{cite}
\fi

\ifCLASSINFOpdf
  \usepackage[pdftex]{graphicx}
  \graphicspath{{./images/}}
\else
\fi
\usepackage{amsmath}

\usepackage{algorithmic}
\usepackage{algorithm}
\usepackage{multirow}
\usepackage[caption=false,font=footnotesize,labelfont=sf,textfont=sf]{subfig}
\begin{document}
%
\title{PERF: Performant, Explicit Radiance Fields}

\author{\IEEEauthorblockN{Sverker Rasmuson}
\IEEEauthorblockA{Chalmers University of Technology\\
Gothenburg, Sweden \\
Email: sverker.rasmuson@chalmers.se}
\and
\IEEEauthorblockN{Erik Sintorn}
\IEEEauthorblockA{Chalmers University of Technology\\
Gothenburg, Sweden\\
Email: erik.sintorn@chalmers.se}
\and
\IEEEauthorblockN{Ulf Assarsson}
\IEEEauthorblockA{Chalmers University of Technology\\
Gothenburg, Sweden\\
Email: uffe@chalmers.se}}

\maketitle

\begin{abstract}
We present a novel way of approaching image-based 3D reconstruction based on radiance fields. The problem of 
volumetric reconstruction is formulated as a non-linear least-squares problem and solved explicitly without the use of
neural networks. This enables the use of solvers with a higher rate of convergence than what is typically used for 
neural networks, and fewer iterations are required until convergence. \\
The volume is represented using a grid of voxels, with the scene surrounded by a hierarchy of environment maps. This 
makes it possible to get clean reconstructions of 360\textdegree\ scenes where the foreground and background is 
separated. \\
A number of synthetic and real scenes from well known benchmark-suites are successfully reconstructed with quality on 
par with state-of-the-art methods, but at significantly reduced reconstruction times.
\end{abstract}

\IEEEpeerreviewmaketitle

\section{Introduction}
We present Performant Explicit Radiance Fields (PERF), a novel approach to image-based volumetric reconstruction of 
360\textdegree\ scenes based on radiance fields. Unlike previous methods, our approach does not use an MLP solution, 
but instead solves the problem directly as a non-linear least-squares system. This system is minimized using a 
Gauss-Newton solver with higher rate of convergence than standard Stochastic Gradient Descent used for systems based 
on Neural Networks. As a result, the time until convergence for a given scene is significantly shorter than for 
comparable methods.

\begin{figure*}[!t]
\begin{minipage}{0.6\linewidth}
\begin{minipage}{0.1\linewidth}
\vfill
\textbf{NERF}
\vfill
\end{minipage}
\begin{minipage}{0.9\linewidth}
\subfloat{\includegraphics[width=0.24\linewidth]{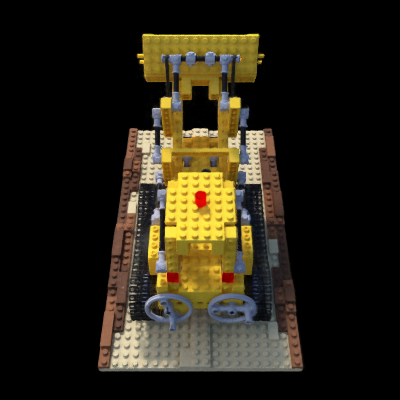}}
\subfloat{\includegraphics[width=0.24\linewidth]{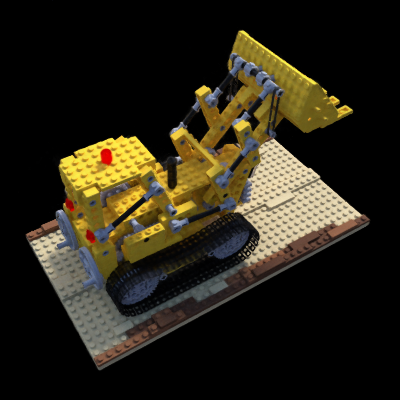}}
\subfloat{\includegraphics[width=0.24\linewidth]{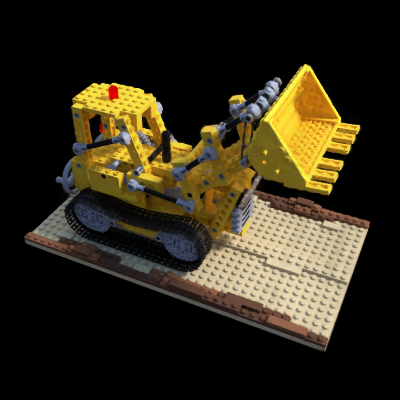}}
\subfloat{\includegraphics[width=0.24\linewidth]{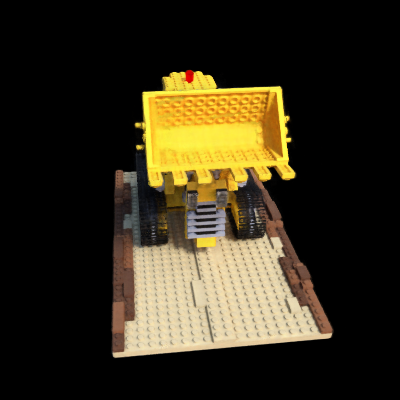}}
\end{minipage}
\begin{minipage}{0.1\linewidth}
\vfill
\textbf{ADAM}
\vfill
\end{minipage}
\begin{minipage}{0.9\linewidth}
\subfloat{\includegraphics[width=0.24\linewidth]{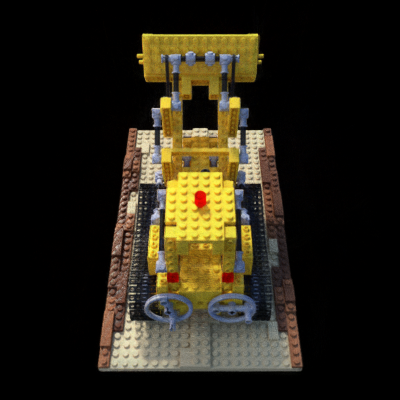}}
\subfloat{\includegraphics[width=0.24\linewidth]{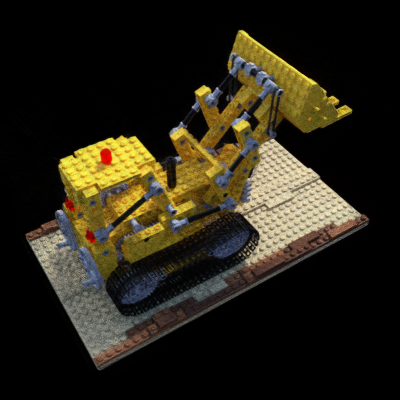}}
\subfloat{\includegraphics[width=0.24\linewidth]{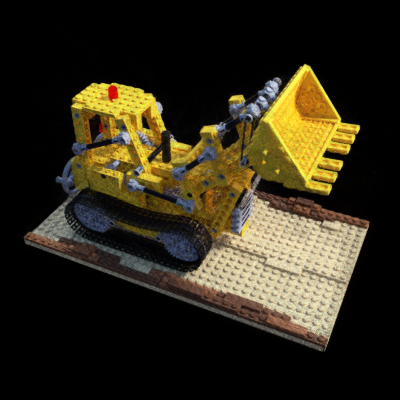}}
\subfloat{\includegraphics[width=0.24\linewidth]{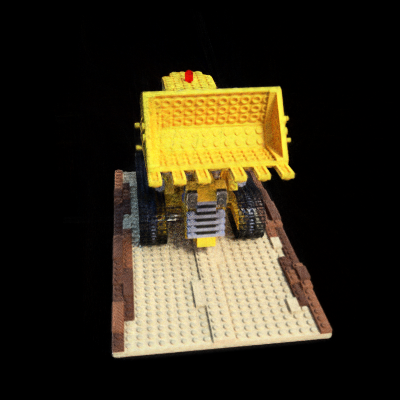}}
\end{minipage}
\begin{minipage}{0.1\linewidth}
\vfill
\textbf{GN}
\vfill
\end{minipage}
\begin{minipage}{0.9\linewidth}
\subfloat{\includegraphics[width=0.24\linewidth]{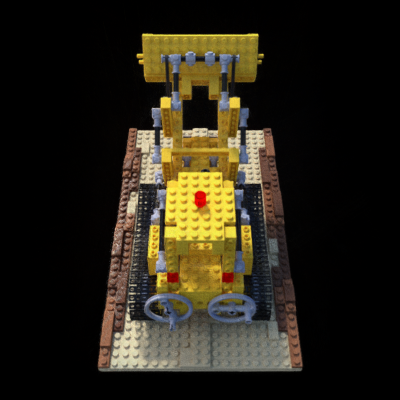}}
\subfloat{\includegraphics[width=0.24\linewidth]{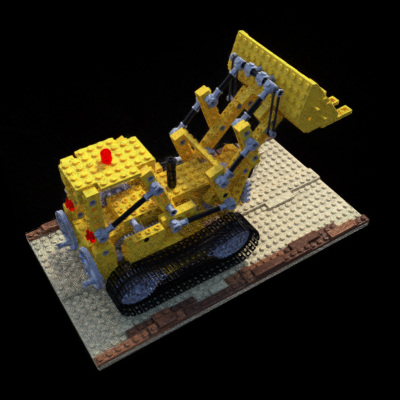}}
\subfloat{\includegraphics[width=0.24\linewidth]{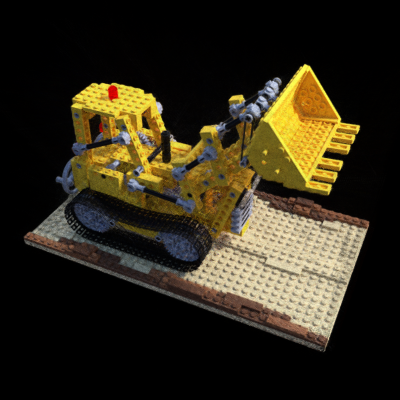}}
\subfloat{\includegraphics[width=0.24\linewidth]{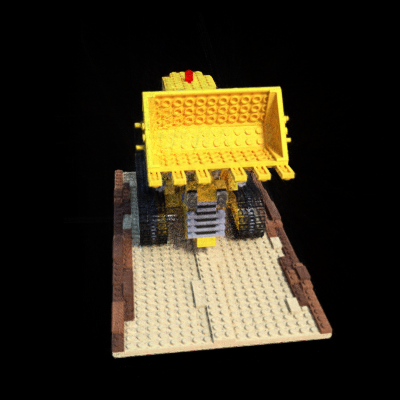}}
\end{minipage}
\begin{minipage}{0.1\linewidth}
\vfill
\textbf{GN-H}
\vfill
\end{minipage}
\begin{minipage}{0.9\linewidth}
\subfloat{\includegraphics[width=0.24\linewidth]{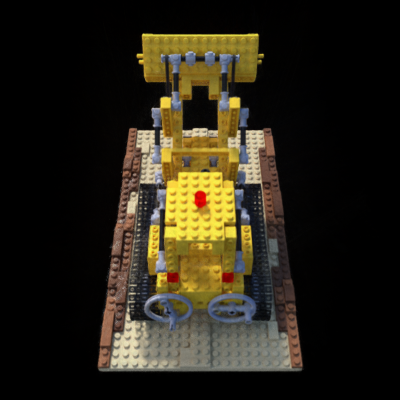}}
\subfloat{\includegraphics[width=0.24\linewidth]{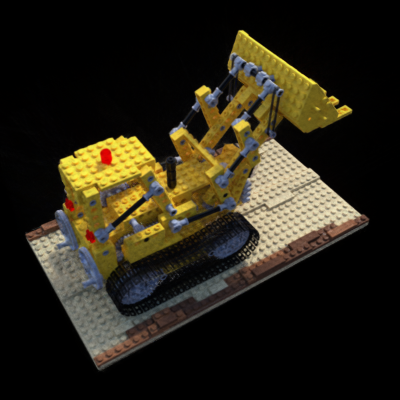}}
\subfloat{\includegraphics[width=0.24\linewidth]{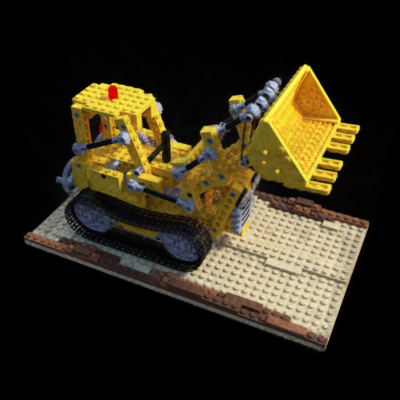}}
\subfloat{\includegraphics[width=0.24\linewidth]{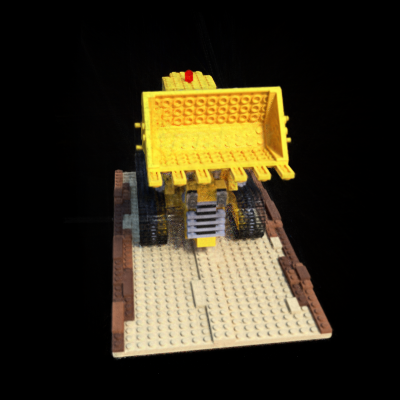}}
\end{minipage}
\end{minipage}
\begin{minipage}{0.3\linewidth}
\subfloat{\includegraphics[height=4cm]{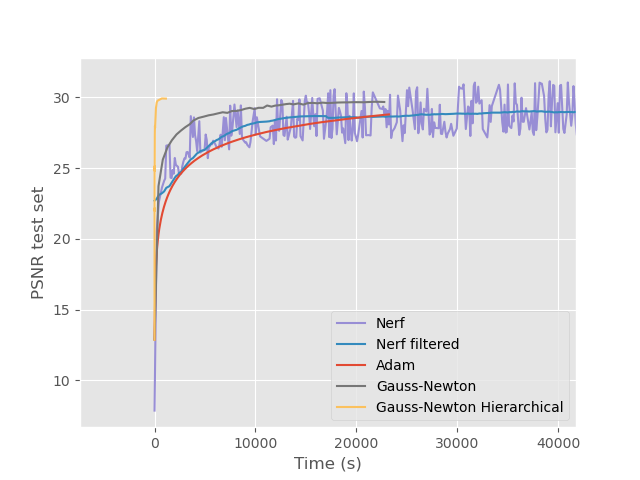}} \\
\subfloat{\includegraphics[height=4cm]{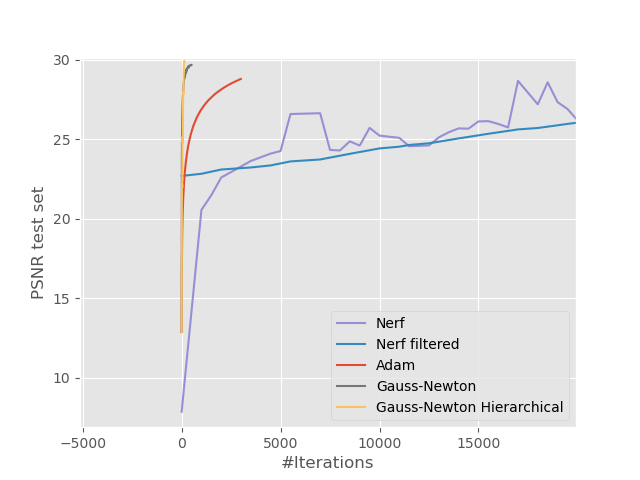}}
\end{minipage}
\caption{Hold out views for the lego scene comparing NERF (without view-dependance), pure ADAM-optimization, regular 
Gauss-Newton optimization and a hierarchical Gauss-Newton formulation; from top to bottom. To the top-right the corresponding 
PSNR-curves are shown versus time; to the bottom-right PSNR is plotted versus number of iterations.}
\label{fig:adam_vs_nerf}
\end{figure*}

The NERF~\cite{Mildenhall2020} techniques present a simple formulation of the 3D reconstruction problem using volume rendering and neural networks. We ask the question if it is possible 
to solve this problem directly using pure ADAM-optimization (see 
Figure~\ref{fig:adam_vs_nerf})~\cite{ADAM2014}. With this confirmed, and identifying the problem as non-linear least-squares, we can see that
it is possible to achieve much higher convergence rates for the same volumetric formulation, without sacrificing much in
terms of quality.

The scene is represented using a grid of voxels containing color and opacity values. Additionally, surrounding the object of interest, is a hierarchical layer of environment maps (also with color and opacity), which enable efficient segmentation of foreground from background.

A ray is followed through each pixel of a given image, and color accumulated through the voxel grid and the 
surrounding environment maps according to classic volumetric rendering. The square of the difference between the 
accumulated ray and the value of the pixel is then computed for each pixel in each image. The sum of all these 
differences builds up to a non-linear least-squares system. In this system, each ray corresponds to three residuals 
(one for each color channel), and each voxel in the grid and texel in the environment maps corresponds to four 
variables each (three color channels and opacity).

This non-linear least squares system can be solved efficiently using the Gauss-Newton method. The Gauss-Newton method 
has several useful properties in being amenable to parallelization, while only requiring the computation of 
partial derivatives of the first degree. We show that these derivatives can be computed analytically with a simple 
iterative algorithm. Each Gauss-Newton step is computed using the iterative Preconditioned Conjugate Gradient 
algorithm. 

Since rendering the volume only amounts to raymarching through the voxel grid and environment maps, inspecting and 
evaluating the reconstruction is possible in real-time during any stage of the computation.

To summarize, the contributions in this paper are:
\begin{itemize}
\item a solution of volumetric reconstruction based on a non-linear least squares formulation,
\item efficient computation of analytical derivatives using an iterative algorithm,
\item a fast GPU-based Gauss-Newton solver using Preconditioned Conjugate Gradient,
\item the separation of foreground and background for 360\textdegree\ scenes using a hierarchical envelope of environment maps.
\end{itemize}

\section{Related work}
\paragraph{\textbf{Multi-view Stereo}}
Multi-view stereo (MVS) is a classical application of computer vision where a scene is reconstructed from
a set of images taken from multiple viewpoints, typically using pairwise stereo matching algorithms.
This is a well-researched topic that have been extensively treated the last few decades. For an overview,
see Hartley and Zisserman~\cite{Zisserman2004} and Seitz
et al.~\cite{seitz2006comparison}. In recent years, this approach have been successfully paired with neural networks
in deep multi-view stereo~\cite{yao2018}\cite{yao2019}.

\paragraph{\textbf{Structure From Motion}}
Another classical approach for scene reconstruction is Structure from Motion
(SfM)~\cite{snavely2006photo}\cite{ozyecsil2017}. In these systems images are paired by identifying
image features, which works well given that there is enough overlap between the images. 
Camera poses (or \textit{motion}) are estimated and 3D points corresponding to image features are computed.
The output of such a reconstruction is sparse, but it can be paired with stereo matching techniques
such as PatchMatch Stereo to create dense output~\cite{Bleyer2011}\cite{schonberger2016structure}.

\paragraph{\textbf{View Synthesis}}
View synthesis is a related research area where focus lies on synthesizing novel viewpoints given a set of images and their poses, not necessarily
caring about the underlying geometry of the scene. This includes image-based rendering and lightfields, which have been explored thoroughly over the years~\cite{shum2000}.

One recent attempt at this problem is to use Multi-Plane Images (MPIs); a set of semi-transparent images that can be trained efficiently with neural networks and
then rendered from a novel viewpoint~\cite{Mildenhall2019}\cite{Zhou2018}.

Broxton et al. surround the scene with a hierarchical set of environment maps in a similar way to ours using MPIs, 
with the goal of representing a scene captured by a fixed rig of 46 synchronized action cameras ~\cite{Broxton2020}.
They, however, use MPIs as the primary data structure to represent their scenes, while in our work we use the surrounding environment maps as a tool to segment the background from the 
region of
interest modeled with voxels. Our work does also not use neural networks to train MPIs, but instead use a layer of 
environment maps optimized explicitly in the same manner (and 
simultaneously) as the voxel grid.

\paragraph{\textbf{Neural Rendering}}
Recently a popular approach have been to use neural networks as a key component in encoding the scene representation.
In Neural Volumes~\cite{Lombardi2019}, a semi-transparent voxel grid is coupled with a encoder-decoder network to create high quality reconstructions from a multi-view setup. While the 
scene is not explicitly stored in a network, they are still pivotal to its overall approach.
In NERF~\cite{Mildenhall2020}, the outgoing radiance $f(\mathbf{x}, \mathbf{\omega})$ at any point, $\mathbf{x}$, and direction, $\mathbf{\omega}$, is expressed as an MLP with a 
positional encoding for the inputs. The positional encoding has recently been shown to improve the ability of an MLP to reconstruct high frequency signals~\cite{Tancik2020a}.
Many improvements have been presented to the original NERF implementation to help with e.g. improved view synthesis and real-time rendering~\cite{martinbrualla2020}\cite{liu2020}\cite{yu2021}.

For an overview of this growing field, we refer the reader to the STAR-paper by Tewari et al.~\cite{Tewari2020NeuralSTAR}.

To our knowledge, only a few papers have tried to address one of the main caveats of the otherwise compelling approach used in NERF; namely the long reconstruction times.

Tancik et al. have demonstrated that, when the target function belongs to a known class, the network can be pre-initialized to make the final 
optimization faster~\cite{Tancik2020}. The authors show that when pretraining on a class of objects in the ShapeNet~\cite{chang2015shapenet} dataset, the result after a few iterations is 
much better than, but the quality converges to the same value as for, a randomly initialized network after as many iterations. An important observation in our paper is that a discrete 
Radiance Field can be obtained much faster by optimizing the function, $f(\mathbf{x}, \mathbf{\omega})$, directly rather than training an MLP representation. 

MVSNerf~\cite{Chen2021} is an alternative approach to creating a Radiance Field, in which a few input views are processed through a CNN to create a feature vector that is projected into a
cost volume defined by a reference camera. These feature vectors are passed, along with position and direction, to an MLP which is optimized as in previous work. This allows for much 
faster training of the network, but unfortunately only for the geometry directly seen by the reference camera. 

In this work, we significantly improve the reconstruction times by identifying the problem of volumetric reconstruction as a non-linear least-squares problem, which can be solved directly
with an efficient Gauss-Newton solver.

\section{Method}
\begin{figure}
\label{fig:env_maps}
\includegraphics[scale=0.1]{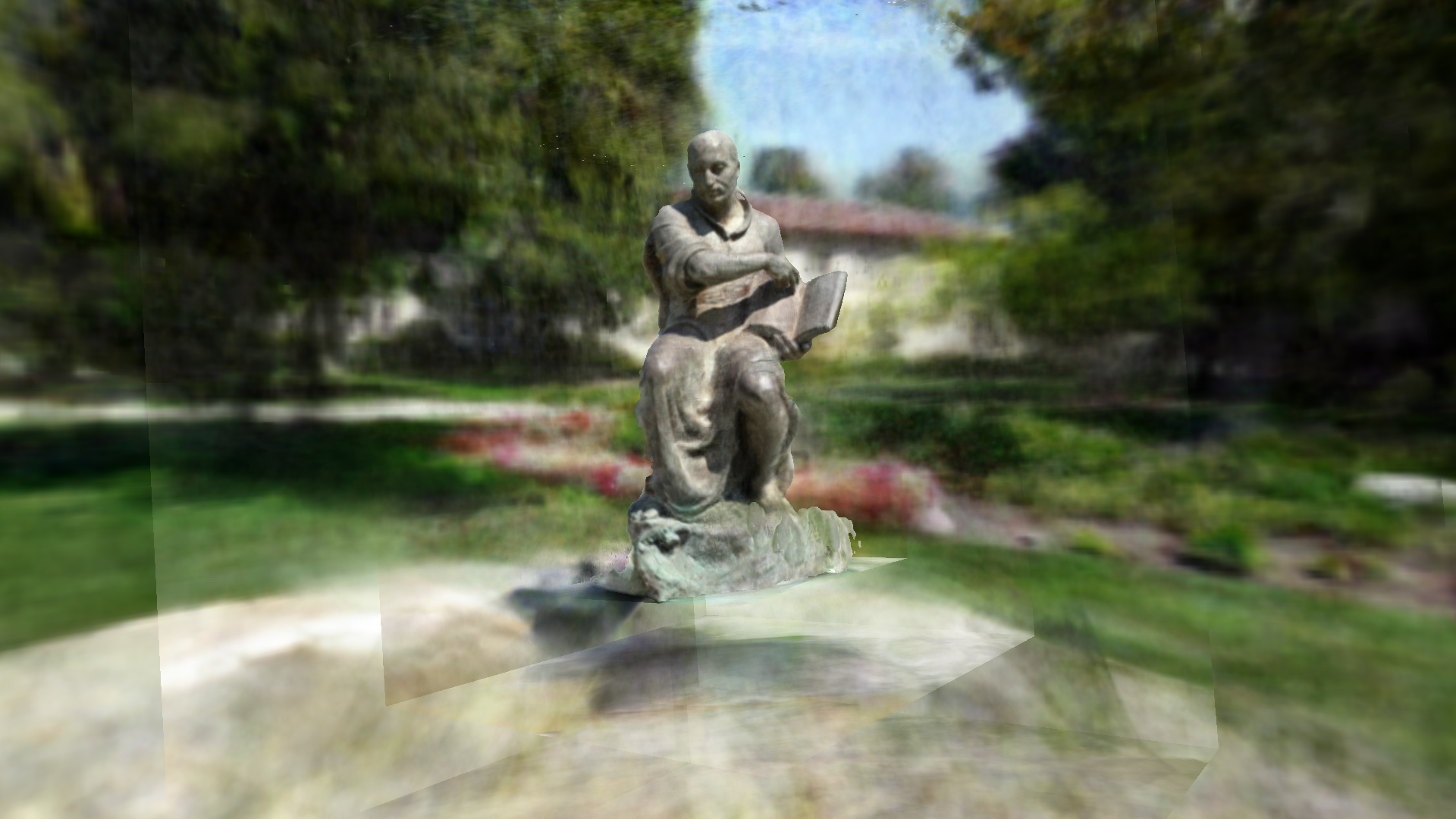}
\caption{Environment maps surround the scene, improving convergence and allowing for separation of foreground and background.}
\end{figure}

We represent the scene using a voxel grid, where each cell contains three color channels, $\mathbf{C} = 
\{c_r, c_g, c_b\}$, and differential
opacity ($\sigma$ in Equation~1-4). The voxel grid is surrounded by 
a hierarchical set of environment maps, centered around the object of interest. These environment maps 
also contain a color and opacity per texel. The main purpose of these environment maps is not to try to 
reconstruct the background, but rather to improve convergence and allow for transparent voxels (see 
Figure~\ref{fig:env_maps}).

\subsection{Volume Rendering}
The render an image, the color of each pixel is computed by integrating over the ray going through the 
pixel and into the volume using the volume rendering equation:
\begin{equation}
    H(\mathbf{r}) = \int_{t_n}^{t_f} T(t)\sigma(\mathbf{r}(t))\mathbf{c}(\mathbf{r}(t))dt
\end{equation}
where
\begin{equation}
   T(t) = e^{(-\int_{t_n}^{t}\sigma(\mathbf{r}(s))ds)}.
\end{equation}

Given that the volume is discretized in a grid, the equation amounts to ray-marching through this grid
according to:
\begin{equation}
    \hat{\mathbf{H}} = \sum_{t_n}^{t_f} \hat{T}(t)(1-e^{(-\sigma(t)\delta(t))})\mathbf{c}(t)
    \label{eq:qr}
\end{equation}
where
\begin{equation}
    \label{eq:alpha}
   \hat{T}(t) = e^{(-\sum_{t_n}^{t}\sigma(t)\delta(t))}.
\end{equation}

\begin{figure}
\label{fig:scene}
\includegraphics[scale=0.5]{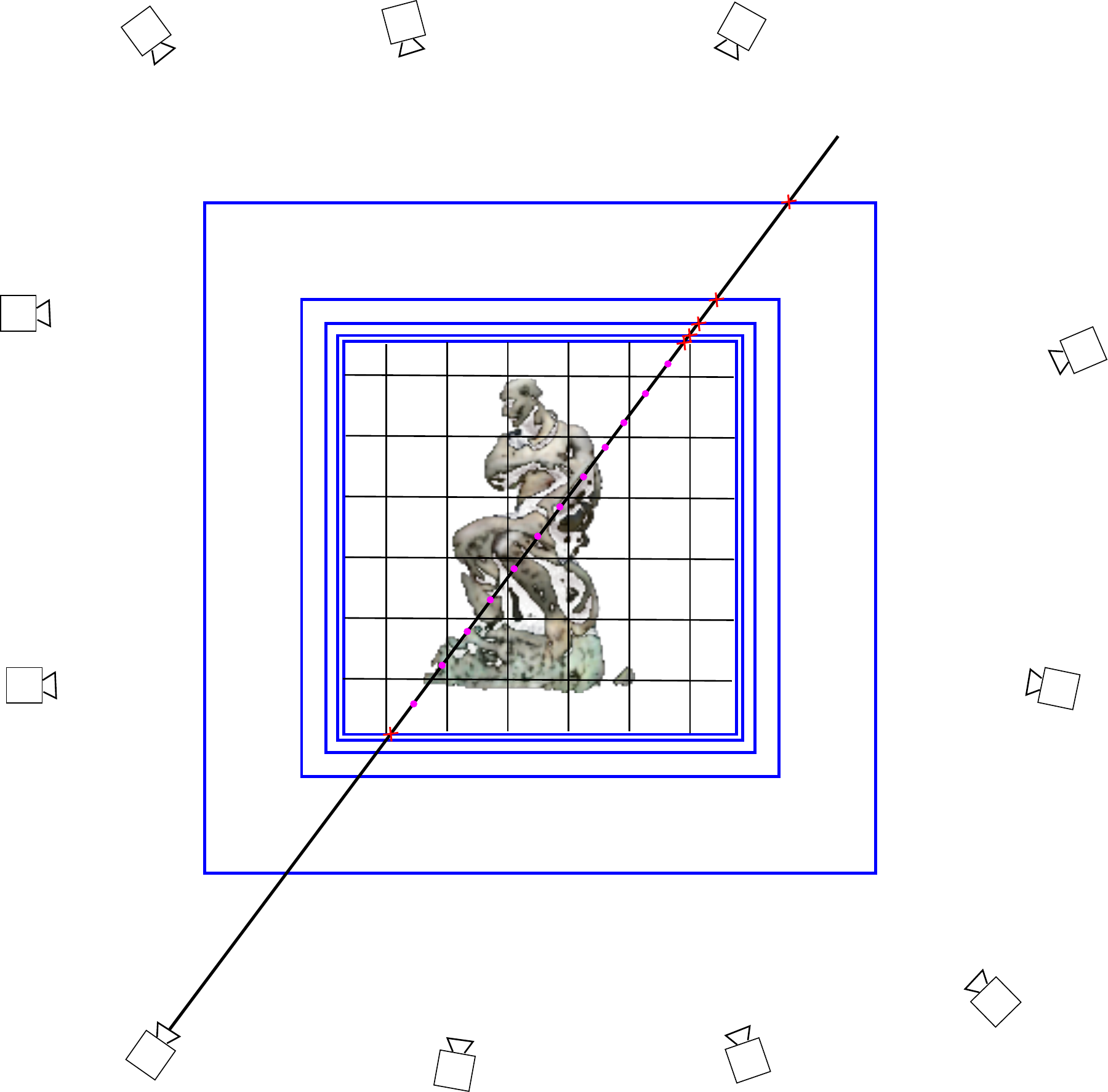}
\caption{A sketch of how a ray is integrated through the scene representation.
The main region of interest is surrounded by a voxel grid. This grid 
is hierarchichally surrounded by environment maps, their distance increasing with the square of the distance to the 
voxel grid. First, an intersection test is made between the ray and the grid to get the entry and exit points. While 
in the volume, the ray is raymarched using jittered steps. Upon exit, the ray is successively intersected with the 
environment maps in order. Intersection tests are marked with an x and raymarch steps with a dot.}
\end{figure}

The contribution from the environment maps are accumulated much in the same manner, but instead of 
ray-marching intersection testing is used, starting from where the ray exists the volume and going 
outwards, see Section~\ref{sec:rend} and Figure~\ref{fig:scene}.

\subsection{Volumetric Reconstruction}
Given a set of images of a scene, we wish to reconstruct the volume enclosed by the corresponding 
cameras in 3D.
Using the formulation of volume rendering from Equation~\ref{eq:qr}, we can formulate the reconstruction 
as an optimization problem that minimizes the sum of squares of the difference between the accumulated 
color for a ray sent through each pixel of every image, and the actual color sampled from the same pixel.
The objective function that we want to minimize is:
\begin{equation}
    F(\mathbf{c}(t), \mathbf{\sigma}(t)) = 0.5\sum_{t_n}^{t_f}\sum_{i = 1}^3(\hat{H}_i(c_i(t),\mathbf{\sigma}(t)) - C_i)^2,
    \label{eq:obj}
\end{equation}

with $\hat{H}_i(c_i(t),\mathbf{\sigma}(t))$ from Equation~\ref{eq:qr} and each pixel color $C_i$.

\subsection{Non-linear least-squares formulation}
The objective function in Equation~\ref{eq:obj} is a non-linear least-squares equation, of which the general form is
\begin{equation}
    S(\mathbf{x}) = 0.5\sum_{i=1}^{m}r_i(\mathbf{x})^2,
    \label{eq:nlls}
\end{equation}

where $r_i$ are called \textit{residuals}.

The problem of volumetric reconstruction can thus be formulated as the minimization of a non-linear 
least-squares problem of this form, where the residuals $r_i$ correspond to the the difference between 
pixel color $\mathbf{C}$ and accumulated colors $\hat{\mathbf{H}}$ for each ray. For each pixel the three color channels are added independently to the sum, giving three residuals per ray.

\subsection{Gauss-Newton solver}
The Gauss-Newton algorithm can be used to iteratively solve non-linear least-squares problems of the form in 
Equation~\ref{eq:nlls}. The Gauss-Newton method can be seen as an approximation of the full Newton's method that works well 
for non-linear least-squares problems. Compared to Newton's method, only first order derivatives need to be computed, while 
it still retains better convergence than simpler algorithms such as Gradient Descent~\cite{Nocedal:2006}. \\
Starting with a value $\mathbf{x}_0$, the update function for the algorithm is:

\begin{equation}
    \mathbf{x}_{i+1} = \mathbf{x}_i - (\mathbf{J_r}^T\mathbf{J_r})^{-1}\mathbf{J_r}^T\mathbf{r}(\mathbf{x}_i)
    \label{eq:gn}
\end{equation}

where $\mathbf{x}_i$ represent all color and opacity values, and where the jacobian $\mathbf{J_r}$ has the structure

\[
\mathbf{J_r} = 
\begin{bmatrix}
\frac{\partial \mathbf{r}_1}{\partial x_1} & \dots & \frac{\partial \mathbf{r}_1}{\partial x_n} \\
\vdots                            &       & \vdots                            \\
\frac{\partial \mathbf{r}_m}{\partial x_1} & \dots & \frac{\partial \mathbf{r}_m}{\partial x_n}
\end{bmatrix}.
\]

Note that each row of $\mathbf{J_r}$ corresponds to the gradient of the residual $\nabla\mathbf{r}_i$.
The matrix $\mathbf{J_r}$ is very sparse, since only a fraction of all $\mathbf{x}_i$'s (corresponding to 
a voxel or texel) will actually be hit by a given ray, and thus be non-zero.

\subsection{Computation of partial derivatives}
For each ray that is marched through the volume, the partial derivatives need to be computed with respect 
to each variable in the voxels or texels that are encountered for Equation~\ref{eq:qr}. We can rewrite 
Equation~\ref{eq:qr} as:

\begin{align*}
    \hat{\mathbf{H}} &=                   e^0(1 - e^{-\sigma_1\delta_1})\mathbf{c}_1  \\
                     &+  e^{-\sigma_1\delta_1}(1 - e^{-\sigma_2\delta_2})\mathbf{c}_2  \\
  &+ e^{-(\sigma_1\delta_1 + \sigma_2\delta_2)}(1 - e^{-\sigma_3\delta_3})\mathbf{c}_3  \\
  &+ \dots                                                                              \\
  &+ e^{-\sum_1^{N-1}\sigma_i\delta_i}(1 - e^{-\sigma_N\delta_N})\mathbf{c}_N
\end{align*}.

The partial derivatives can be efficiently computed with only two auxiliary variables if this equation is traversed backwards (Algorithm~\ref{algo:pd}).

\begin{algorithm}[H]
\caption{Iterative algorithm for computation of partial derivatives of Equation~\ref{eq:qr}.} \label{algo:pd}
\begin{algorithmic}
\STATE $V = e^{-\sum_1^{N}\sigma_i\delta_i}$
\STATE $ \mathbf{G} = \mathbf{0}$
\FOR{i = N \TO i = 1}
\STATE{\begin{align*}
V &= V - \sigma_i\delta_i \\
\frac{\partial \mathbf{H}}{\partial \mathbf{c}_i} &= \mathbf{e^{-V}(1-e^{-\sigma_i\delta_i})} \\
\frac{\partial \mathbf{H}}{\partial \sigma_i} &= e^{-V}\delta_i e^{-\sigma_i\delta_i}\mathbf{c}_i - \delta_i \mathbf{G} \\
\mathbf{G} &= \mathbf{G} + e^{-V}(1 - e^{-\sigma_i\delta_i})\mathbf{c}_i
\end{align*}}
\ENDFOR
\end{algorithmic}
\end{algorithm}

\subsection{Update step}
The update step of Equation~\ref{eq:gn} requires an expensive matrix inversion of the large and sparse 
square matrix $\mathbf{J_r}^T\mathbf{J_r}$. Instead of performing this operation explicitly it is possible
to rewrite the step calculation in the following way for a step $\mathbf{\Delta}$:
\begin{align*}
\mathbf{\Delta} &= - (\mathbf{J_r}^\top\mathbf{J_r})^{-1}\mathbf{J_r}^T\mathbf{r} \\
&\Longleftrightarrow \\
    \mathbf{J_r}^\top\mathbf{J_r}\mathbf{\Delta} &= -\mathbf{J_r}^\top\mathbf{r}.
\end{align*}

With $\mathbf{A} = \mathbf{J_r}^\top\mathbf{J_r}$, $\mathbf{x} = \mathbf{\Delta}$ and $\mathbf{b} = -\mathbf{J_r}^\top\mathbf{r}$ this amounts to a linear system of equations of the standard form
\begin{equation}
\mathbf{A}\mathbf{x} = \mathbf{b}.
\label{eq:lse}
\end{equation}

$\mathbf{A}$ is a very large and sparse matrix ($nxn$ where $n$ is the total number of parameters in the voxel grid and 
environment maps). One efficient way of solving Equation~\ref{eq:lse} is to use the
iterative Preconditioned Conjugate Gradient (PCG) algorithm 
(Algorithm~\ref{algo:pcg}). 

\begin{algorithm}[H]
\caption{The iterative Preconditioned Conjugate Gradient algorithm, with precondition matrix $\mathbf{M}$ (see Subsection~\ref{sec:precon}).}\label{algo:pcg}
\begin{algorithmic}
\STATE $\mathbf{r}_0 = \mathbf{b} - \mathbf{A}\mathbf{x}_0$
\STATE $R_{prev} = \mathbf{r}_{0}^\top\mathbf{r}_{0}$
\STATE $\mathbf{z}_0 = \mathbf{M}^{-1}\mathbf{r}_0$
\STATE $\mathbf{p}_0 = \mathbf{z}_0$
\STATE $k = 0$
\WHILE{true}
\STATE{
\begin{align*}
    \alpha_k &= \frac{\mathbf{r}_k^\top\mathbf{z}_k}{\mathbf{p}_k^\top\mathbf{A}\mathbf{p}_k} \\
    \mathbf{x}_{k+1} &= \mathbf{x}_k + \alpha_k\mathbf{p}_k \\
    \mathbf{r}_{k+1} &= \mathbf{r}_k - \alpha_k\mathbf{A}\mathbf{p}_k \\
    R &= \mathbf{r}_{k+1}^\top\mathbf{r}_{k+1}
\end{align*}
    \IF{$R < EPS$ \textbf{OR} $R / R_{prev} > 0.85$}
    \STATE{\textbf{return} $\mathbf{x}_{k+1}$} 
    \ENDIF
\begin{align*}
    R_{prev} &= R \\
    \mathbf{z}_{k+1} &= \mathbf{M}^{-1}\mathbf{r}_{k+1} \\
    \beta_k &= \frac{\mathbf{r}_{k+1}^\top\mathbf{z}_{k+1}}{\mathbf{r}_k^\top\mathbf{z}_k} \\
    \mathbf{p}_{k+1} &= \mathbf{z}_{k+1} + \beta_k\mathbf{p}_k \\
    k &= k + 1
\end{align*}
}
\ENDWHILE

\end{algorithmic}
\end{algorithm}


The variables $\mathbf{r}$, $\mathbf{z}$, $\mathbf{p}$ and $\mathbf{x}$ are temporary variables with the same size and shape as the voxel grid (or environment maps if they are used).
The vector addition (such as $\mathbf{x}_k + \alpha_k\mathbf{p}_k$)
and dot products (such as $\mathbf{r}_k^\top\mathbf{z}_k$) of algorithm~\ref{algo:pcg} are cheap and easy to compute in parallel on the GPU. The computation of $\mathbf{A}\mathbf{p}$, however, requires some additional exposition. The matrix $\mathbf{A}$ itself is much too large to form explicitly (for a very modest $32^3$ grid it amounts to 4TB of data). To get around this, the following identity can be used:

\begin{equation}
    \mathbf{A}\mathbf{p} = \mathbf{J}_r^\top\mathbf{J}_r\mathbf{p} = \sum_i\nabla\mathbf{r}_i(\nabla\mathbf{r}_i^\top\mathbf{p}).
\end{equation}

This enables each residual to add independently to the computation of $\mathbf{Ap}$, resulting in efficient parallelization. 
For the computation of the gradients $\nabla\mathbf{r}_i$ we use algorithm~\ref{algo:pd}. The algorithm is invoked twice; 
once to compute the sum of the dot product $\nabla\mathbf{r}_i^\top\mathbf{p}$, and once to compute the scalar-vector product 
with this sum and the gradient $\nabla\mathbf{r}_i$. 

\subsection{Preconditioner}\label{sec:precon}
For the precondition matrix $M$ we use a simple diagonal Jacobi preconditioner. This can be computed jointly with the 
gradients since

\begin{equation*}
    diag(\mathbf{J_r}^\top\mathbf{J_r}) =
        \left[\sum(\frac{\partial \mathbf{r}_i}{\partial x_1})^2, \sum(\frac{\partial \mathbf{r}_i}{\partial x_2})^2 , \dots, \sum(\frac{\partial \mathbf{r}_i}{\partial x_n})^2\right].
\end{equation*}

The size of these diagonal elements amounts to a total of $n$ variables, the same as for the voxel grid. This also implies 
that each precondition with $M^{-1}$ simplifies to regular vector addition.

Since each iteration of Algorithm~\ref{algo:pcg} is quite expensive, especially the computation of $\mathbf{Ap}$, we strive to keep the number of iterations of the PCG algorithm low. If the convergence rate is not high enough, then we have found it preferable to exit early (the second condition in algorithm~\ref{algo:pcg}) and continue with the next Gauss-Newton iteration.
Typically, $1-3$ iterations of the PCG algorithm is evaluated.

\subsection{Line search}
In the Gauss-Newton method, there is no guarantee that the objective function will decrease at every 
iteration. It can however be shown that the direction $\mathbf{\Delta} = 
-(\mathbf{J_r}^\top\mathbf{J_r})^{-1}\mathbf{J_r}^T\mathbf{r}$ is a descent direction. Therefore, taking a
a sufficiently small step $\gamma$ in this direction should take us closer to a solution.
In the update function of the Gauss-Newton algorithm (Equation~\ref{eq:gn}), there is no scaling of the
step length. We therefore add a simple backtracking 
algorithm that starts at the initial step length $\gamma = 1$ and scales the step size
by geometrical decimation with a factor of $\alpha = 0.7$. We then use a 
simple heuristic to decide if decimation should continue based on evaluating Equation~\ref{eq:nlls} (see 
Algorithm~\ref{algo:bt}).

\begin{algorithm}[H]
\caption{The backtracking algorithm used to find a suitable step size.}\label{algo:bt}
\begin{algorithmic}
\STATE $\gamma = 1.0$
\STATE $\alpha = 0.7$
\STATE $\mu_{prev} = FLT\_MAX$
\WHILE{true}
\STATE{
$\mu = \text{Error}(\gamma)$     // Equation~\ref{eq:obj} \\ 
\IF {$\mu > \mu_{prev}$} 
\STATE{ \textbf{return} $\gamma/\alpha$}   
\ENDIF \\
$\gamma = \alpha\gamma$ \\
$\mu_{prev} = \mu$
}
\ENDWHILE

\end{algorithmic}
\end{algorithm}

\subsection{Auxiliary terms}\label{sec:aux}
\begin{figure}
\label{fig:flying}
\includegraphics[scale=0.1]{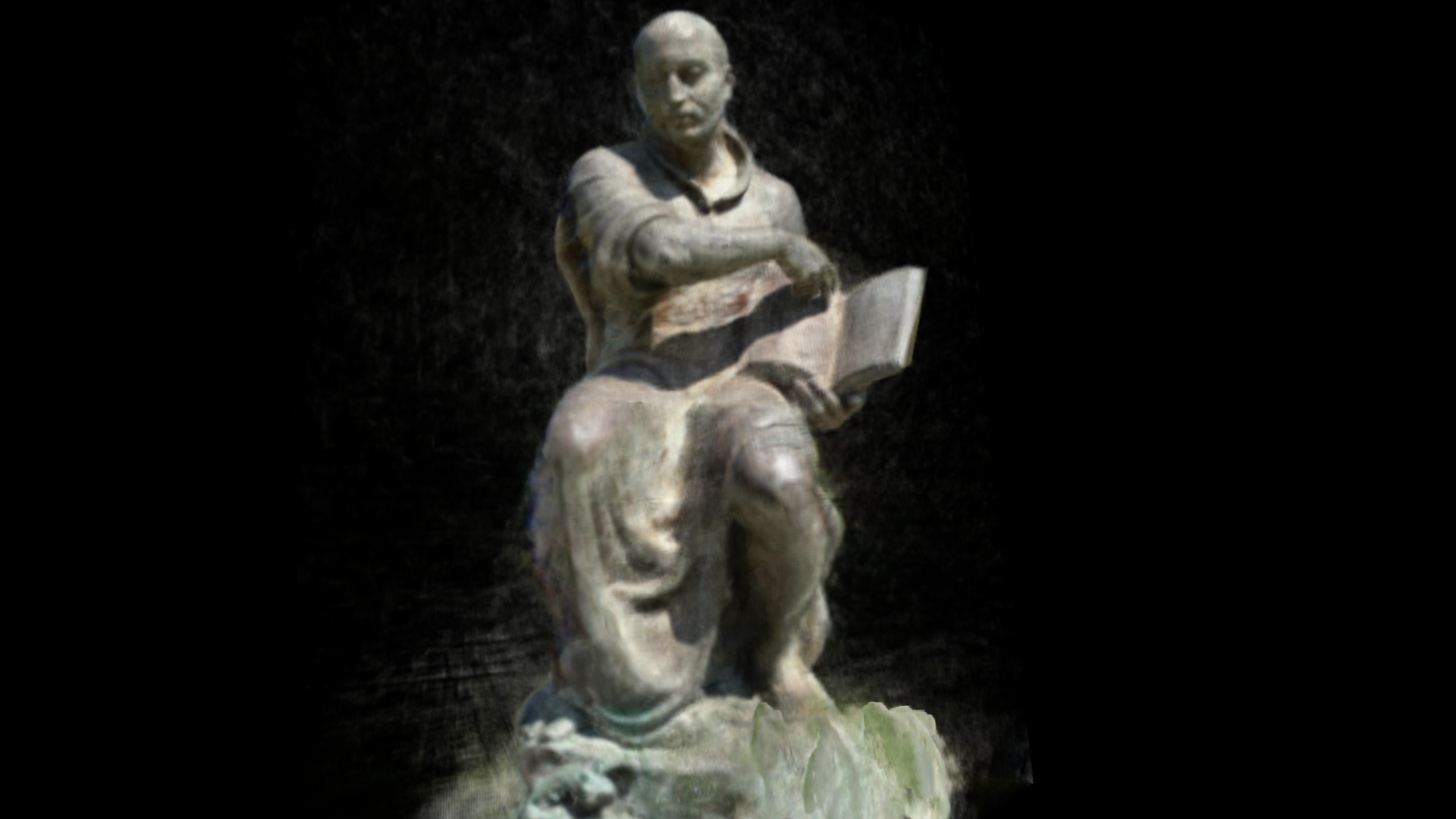}
\caption{Smoke-like artifacts is a way for the algorithm to compensate for view-dependent surfaces.}
\end{figure}

One way for the algorithm to handle view-dependent surfaces is to add highlight-colored smoke-like artifacts in the volume 
(see Figure~\ref{fig:flying}). To mitigate this problem, one extra residual is added per ray. This extra term penalizes the 
total accumulated opacity along a ray if it deviates from 0 or 1. The reasoning is that either the ray should see background 
and thus the path should be fully transparent, or that it hits the object of interest in which case it should be fully opaque.
The following quadratic is used:
\begin{equation}
   L = \lambda(-4(\hat{T} - 0.5)^2 + 1)
\end{equation}

with $\hat{T}$ from Equation~\ref{eq:alpha} and the factor $\lambda = 0.1$.

\section{Implementation}\label{sec:impl}
The Gauss-Newton solver is implemented in CUDA. The grid data is stored in a 3D surface, while the environment maps are stored in cube maps. To facilitate the use of atomics in CUDA, the backing storage is copied to regular memory for some kernels, since this feature is not supported for surfaces. CUDA texture objects are bound to surfaces when sampling, giving access to hardware interpolation.

\subsection{Hierarchical formulation}
To speed up reconstruction further, a simple hierarchical scheme is used, typically with 4-5 hierarchical levels. The resolution in each dimension is doubled for the voxel grid and the 
environment maps for each increasing level. A fixed number of iterations (typically $N = 30$) are performed per level, starting at a resolution of $(32, 32, 32)$ for the voxel grid and 
$(32, 32)$ for each cubemap. The initial values are typically random colors with low opacity. When a level is completed, the tentative result is stored temporarily while all buffers 
are reallocated. The new buffers are then populated with the old data 
using linear interpolation, and computation is resumed at the new hierarchical level. The optimization is sensitive to the initial state of the colors and opacity, and by priming each new
hierarchical level with the result from the previous one, a reasonable start guess is acquired from start. Also, the algorithm only starts from random values for a very low resolution, 
where each iteration is cheap to compute and the extra time spent is negligible.

\subsection{Jittering}
The position within each pixel which the ray originates from is chosen randomly according to two uniformly distributed floats. Jittering is also applied along each ray, randomizing the  
position of which samples are taken within half a raymarch step before and after the uniform steps along the ray.



\subsection{Rendering}\label{sec:rend}
Real-time rendering of a given scene is done with Equation~\ref{eq:qr}. This allows for continuous inspection of a scene during training from any given viewpoint. 
First, an intersection test is made with the ray and bounding box of the voxel grid. The grid is then ray-marched with a step length equal to the shortest side of a voxel cell.
Exiting the volume, the ray is intersected in order with the surrounding environment maps

Upon convergence of a scene, some of flying artifacts are usually still present,  Figure~\ref{fig:flying}. To remove these, we simply add a minimum threshold $(1.0 - T) < 0.7$ for the total opacity along a ray within the target volume when rendering the converged scene. Failing this test, the occupying volume is considered fully transparent.

\section{Results}
We have used our method on a number of different scenes among the NERF data sets, as well as for the Tanks and Temples data set~\cite{Knapitsch2017}. As a reference, we compare our reconstructions with those from NERF with view-dependance disabled to make for a viable comparison. The final hierarchical level of the voxel grid is typically $256^3$ or $512^3$. All results are computed on a single Nvidia Titan V GPU.

\subsection{Reconstruction quality}
\begin{figure*}[!t]
\begin{minipage}{0.6\linewidth}
\begin{minipage}{0.1\linewidth}
\vfill
\textbf{Ours}
\vfill
\end{minipage}
\begin{minipage}{0.9\linewidth}
\subfloat{\includegraphics[width=0.24\linewidth]{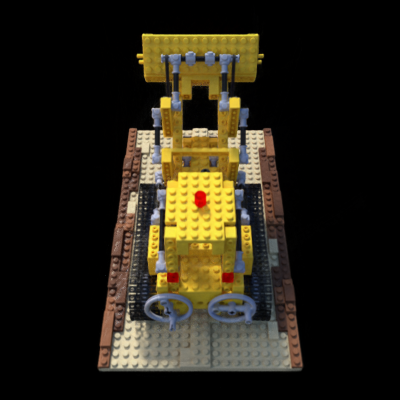}}
\subfloat{\includegraphics[width=0.24\linewidth]{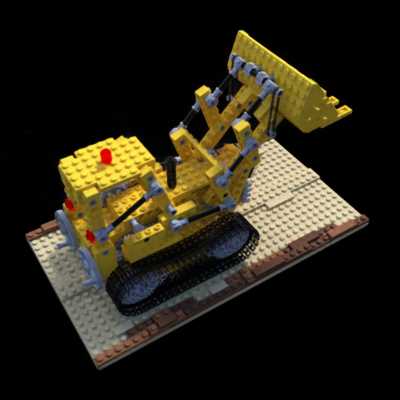}}
\subfloat{\includegraphics[width=0.24\linewidth]{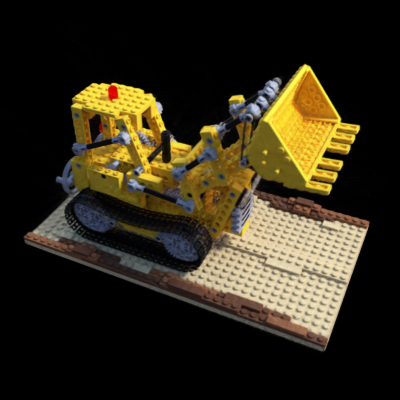}}
\subfloat{\includegraphics[width=0.24\linewidth]{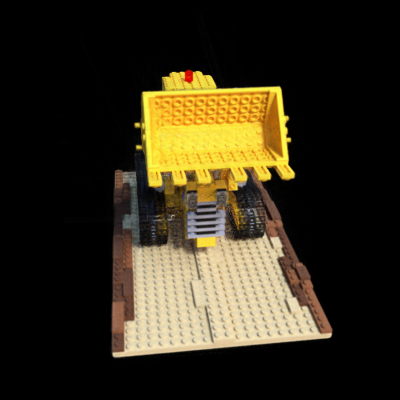}}
\end{minipage}

\begin{minipage}{0.1\linewidth}
\vfill
\textbf{Nerf}
\vfill
\end{minipage}
\begin{minipage}{0.9\linewidth}
\subfloat{\includegraphics[width=0.24\linewidth]{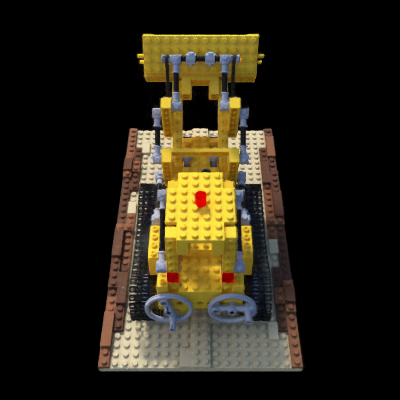}}
\subfloat{\includegraphics[width=0.24\linewidth]{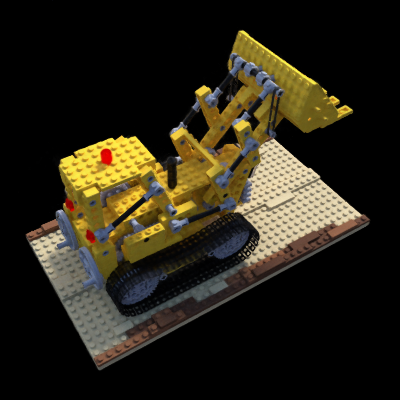}}
\subfloat{\includegraphics[width=0.24\linewidth]{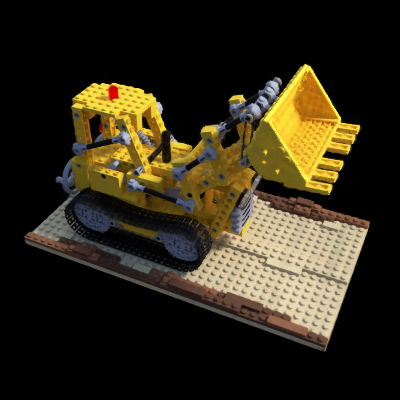}}
\subfloat{\includegraphics[width=0.24\linewidth]{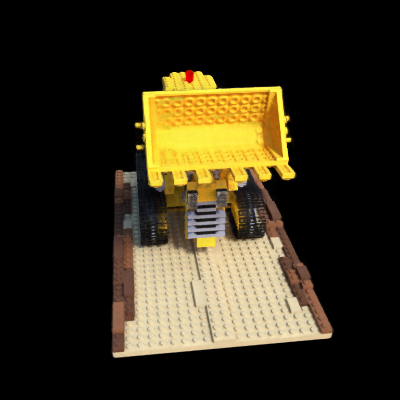}}
\end{minipage}
\end{minipage}
\begin{minipage}{0.4\linewidth}
\subfloat{\includegraphics[width=1.0\linewidth]{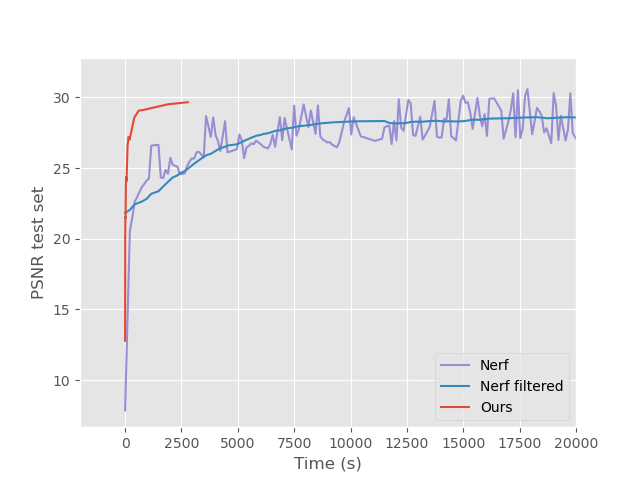}}
\end{minipage}
\begin{minipage}{0.6\linewidth}
\begin{minipage}{0.1\linewidth}
\vfill
\textbf{Ours}
\vfill
\end{minipage}
\begin{minipage}{0.9\linewidth}
\subfloat{\includegraphics[width=0.24\linewidth]{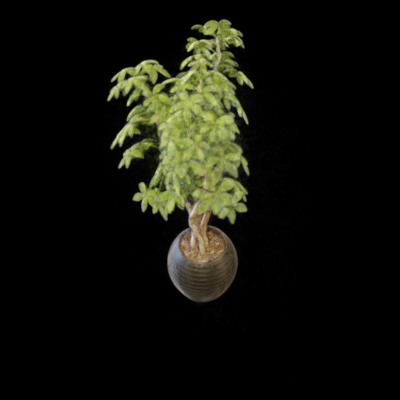}}
\subfloat{\includegraphics[width=0.24\linewidth]{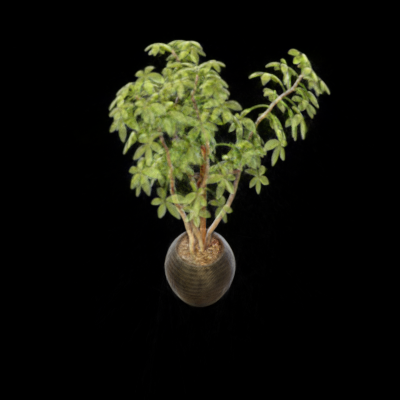}}
\subfloat{\includegraphics[width=0.24\linewidth]{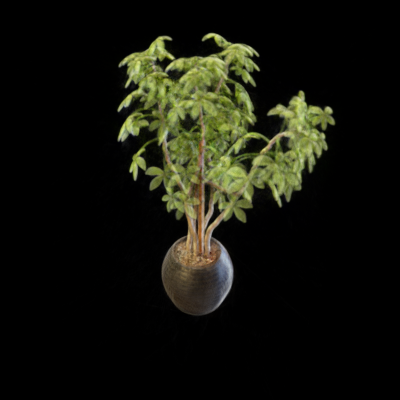}}
\subfloat{\includegraphics[width=0.24\linewidth]{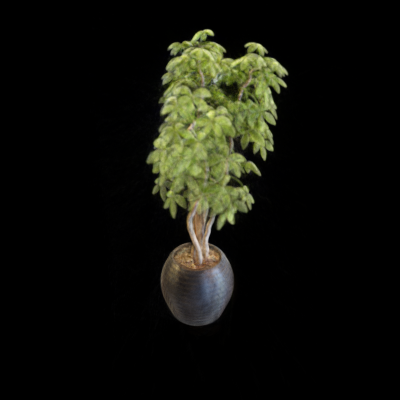}}
\end{minipage}

\begin{minipage}{0.1\linewidth}
\vfill
\textbf{Nerf}
\vfill
\end{minipage}
\begin{minipage}{0.9\linewidth}
\subfloat{\includegraphics[width=0.24\linewidth]{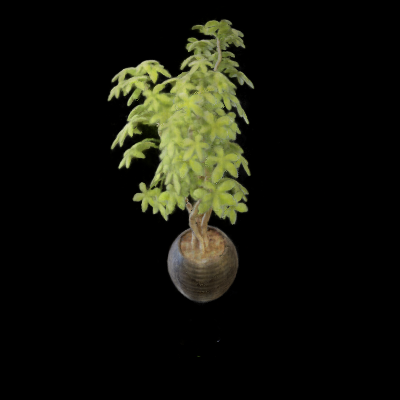}}
\subfloat{\includegraphics[width=0.24\linewidth]{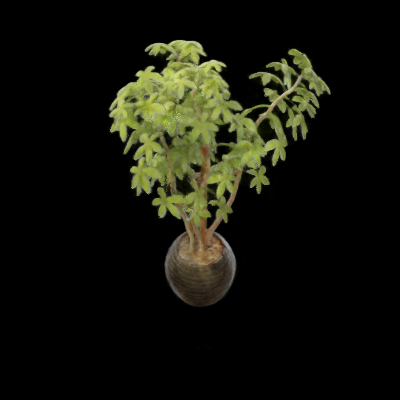}}
\subfloat{\includegraphics[width=0.24\linewidth]{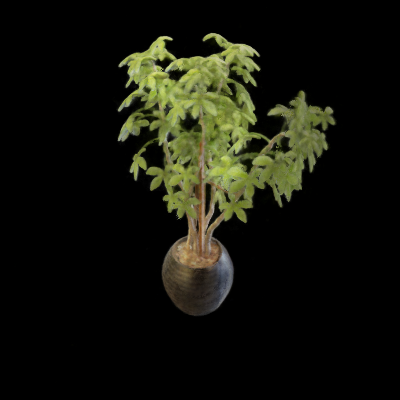}}
\subfloat{\includegraphics[width=0.24\linewidth]{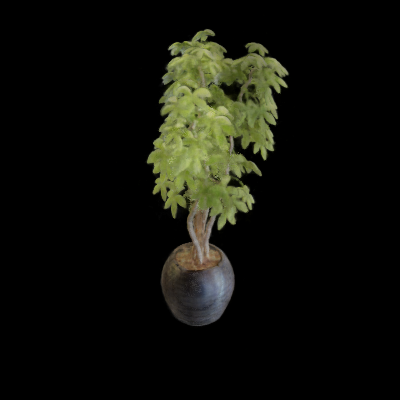}}
\end{minipage}
\end{minipage}
\begin{minipage}{0.4\linewidth}
\subfloat{\includegraphics[width=1.0\linewidth]{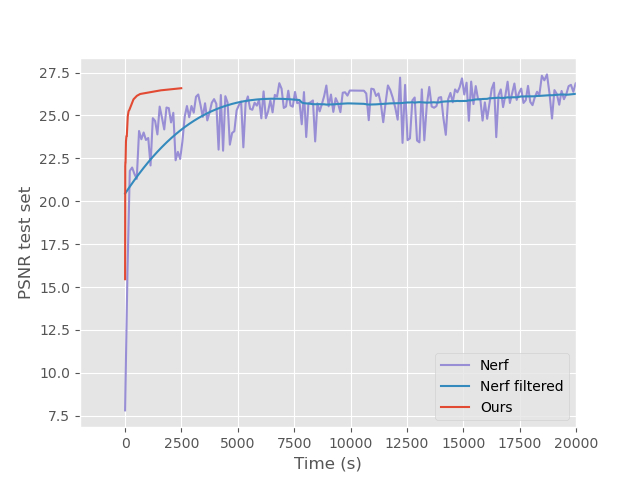}}
\end{minipage}
\begin{minipage}{0.6\linewidth}
\begin{minipage}{0.1\linewidth}
\vfill
\textbf{Ours}
\vfill
\end{minipage}
\begin{minipage}{0.9\linewidth}
\subfloat{\includegraphics[width=0.24\linewidth]{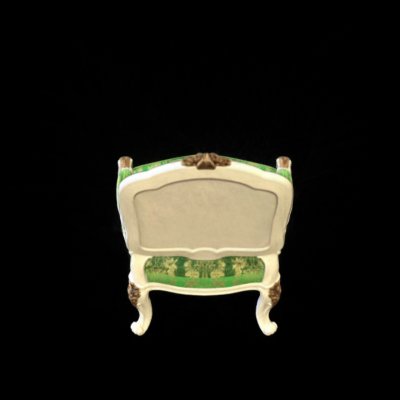}}
\subfloat{\includegraphics[width=0.24\linewidth]{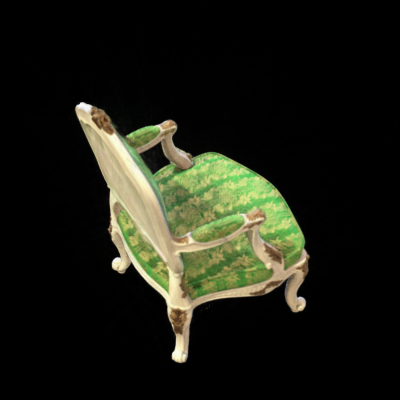}}
\subfloat{\includegraphics[width=0.24\linewidth]{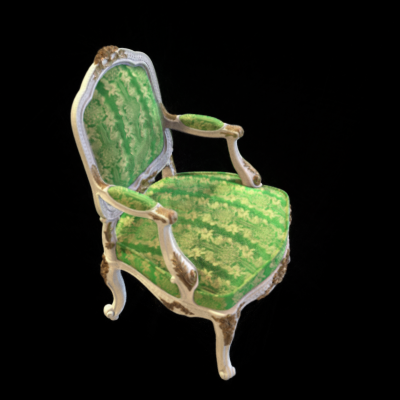}}
\subfloat{\includegraphics[width=0.24\linewidth]{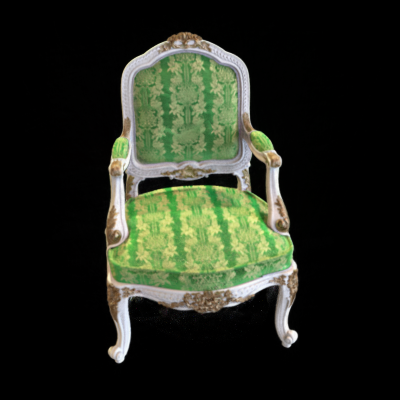}}
\end{minipage}

\begin{minipage}{0.1\linewidth}
\vfill
\textbf{Nerf}
\vfill
\end{minipage}
\begin{minipage}{0.9\linewidth}
\subfloat{\includegraphics[width=0.24\linewidth]{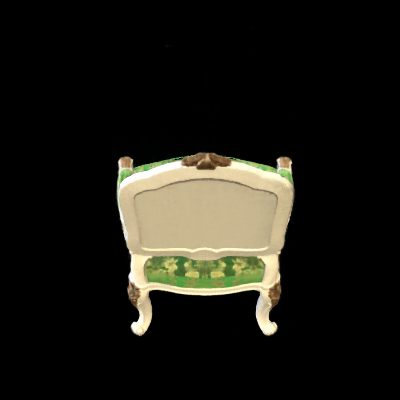}}
\subfloat{\includegraphics[width=0.24\linewidth]{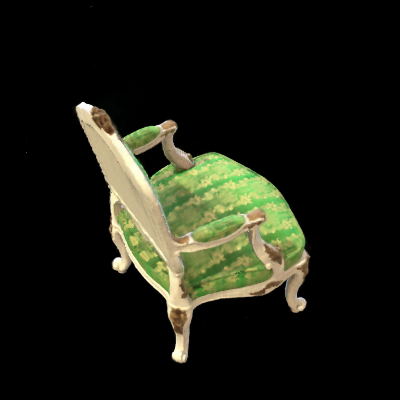}}
\subfloat{\includegraphics[width=0.24\linewidth]{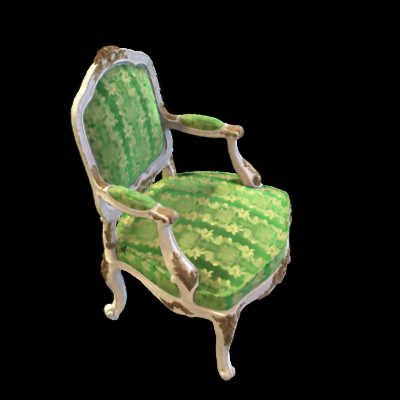}}
\subfloat{\includegraphics[width=0.24\linewidth]{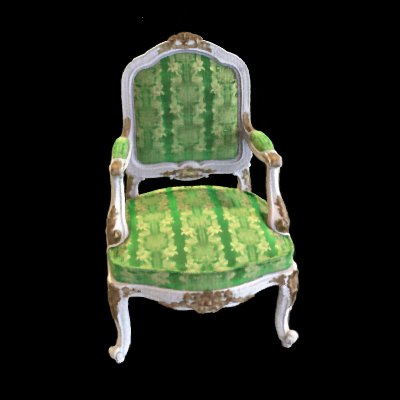}}
\end{minipage}
\end{minipage}
\begin{minipage}{0.4\linewidth}
\subfloat{\includegraphics[width=1.0\linewidth]{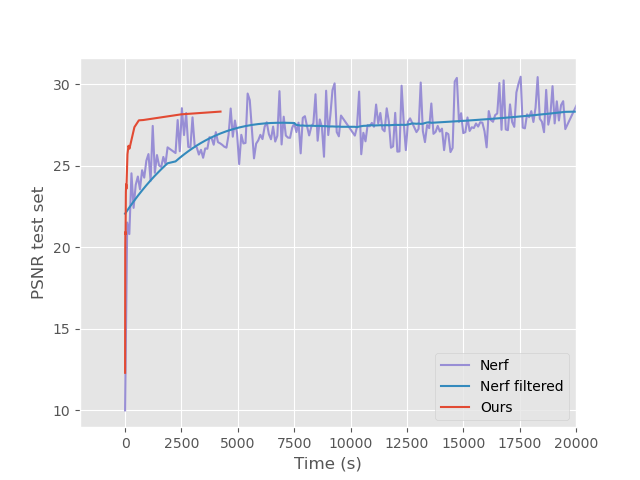}}
\end{minipage}
\begin{minipage}{0.6\linewidth}
\begin{minipage}{0.1\linewidth}
\vfill
\textbf{Ours}
\vfill
\end{minipage}
\begin{minipage}{0.9\linewidth}
\subfloat{\includegraphics[width=0.24\linewidth]{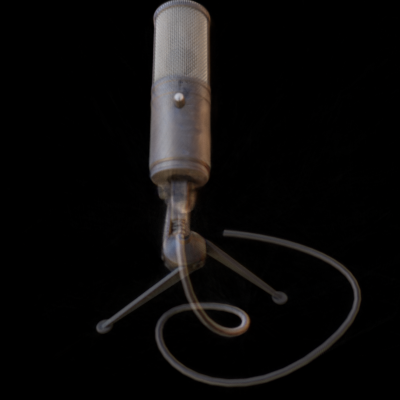}}
\subfloat{\includegraphics[width=0.24\linewidth]{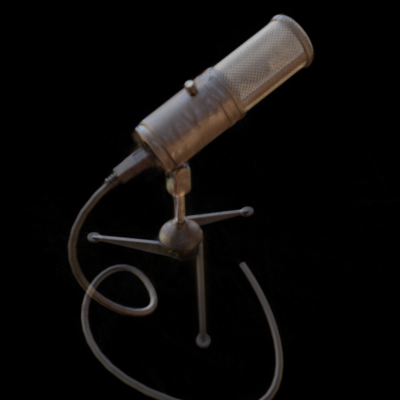}}
\subfloat{\includegraphics[width=0.24\linewidth]{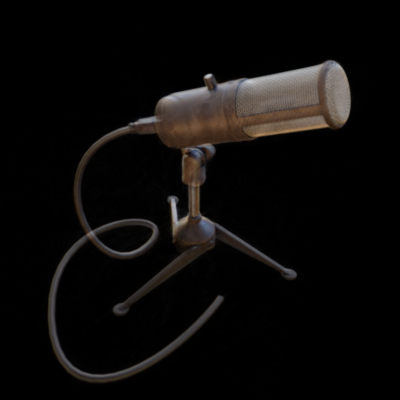}}
\subfloat{\includegraphics[width=0.24\linewidth]{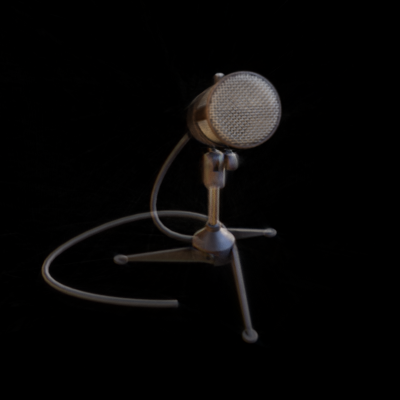}}
\end{minipage}

\begin{minipage}{0.1\linewidth}
\vfill
\textbf{Nerf}
\vfill
\end{minipage}
\begin{minipage}{0.9\linewidth}
\subfloat{\includegraphics[width=0.24\linewidth]{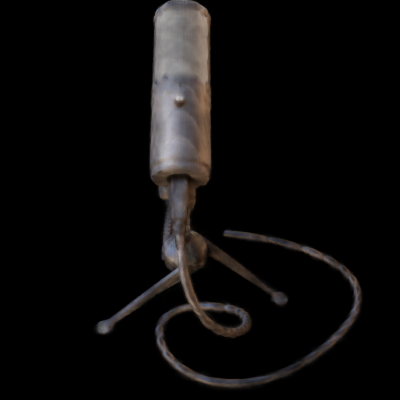}}
\subfloat{\includegraphics[width=0.24\linewidth]{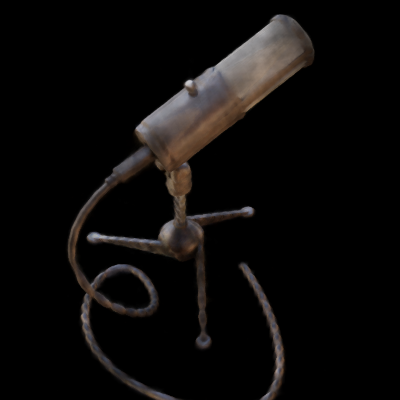}}
\subfloat{\includegraphics[width=0.24\linewidth]{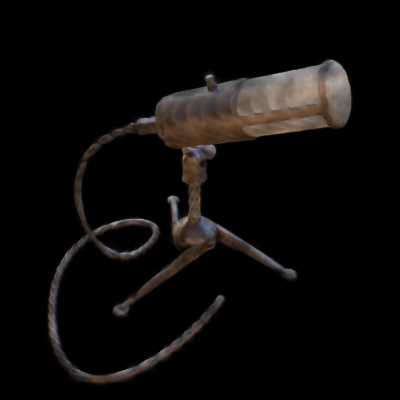}}
\subfloat{\includegraphics[width=0.24\linewidth]{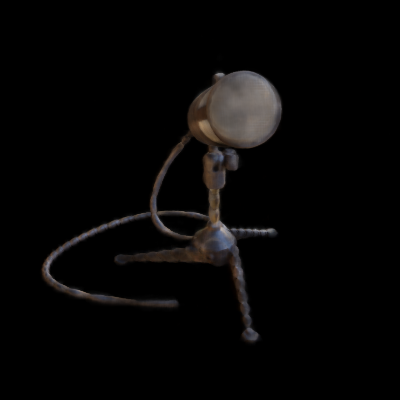}}
\end{minipage}
\end{minipage}
\begin{minipage}{0.4\linewidth}
\subfloat{\includegraphics[width=1.0\linewidth]{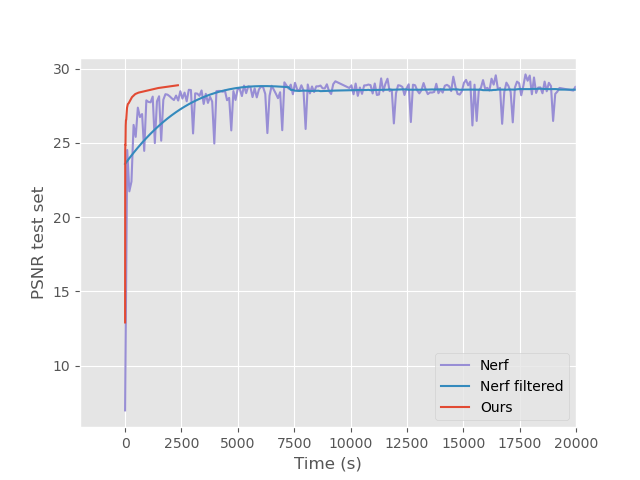}}
\end{minipage}

\caption{Results on the NERF synthetic data set. Our reconstructions are shown for a few hold-out views compared NERF without view-dependance. On the right corresponding curves of PSNR 
versus training time are shown. We can see that our method converges much faster, while obtaining a similar level of quality.}
\label{fig:nerfsyn}
\end{figure*}

In Figure~\ref{fig:nerfsyn}, our method is tested on a number of synthetic scenes from the NERF data set. In all these cases,
we achieve similar or better results compared to the reference. We have purposefully excluded scenes with too much
specularities from these sets, since that is not handled well by either us or NERF without view-dependance.
The bottom microphone scene in Figure~\ref{fig:nerfsyn} is on the limit of what the methods can handle. In this case, our method 
manages to average out specularities in a more pleasing way compared to NERF.

\begin{figure*}[!t]
\begin{minipage}{0.2\linewidth}
\begin{tabular}{cc}
\multicolumn{2}{c}{\textbf{Ours}}                                                \\ \hline
PSNR & Time \\
27.96 & 24m54s
\end{tabular}
\end{minipage}
\begin{minipage}{0.8\linewidth}
\subfloat{\includegraphics[width=0.11\linewidth]{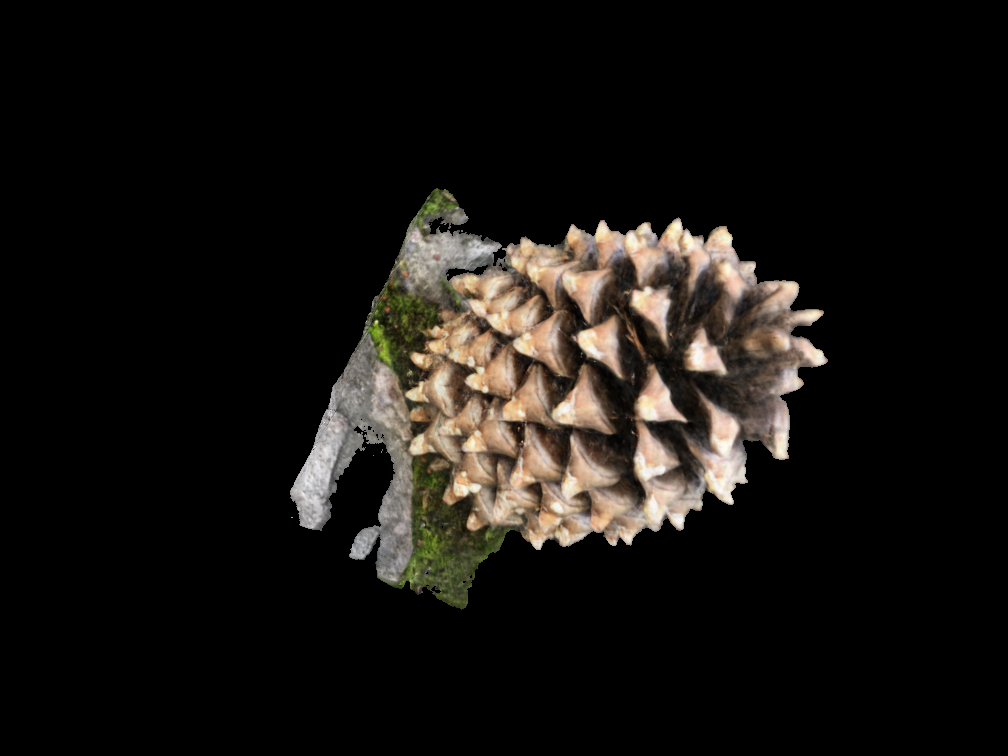}}
\subfloat{\includegraphics[width=0.11\linewidth]{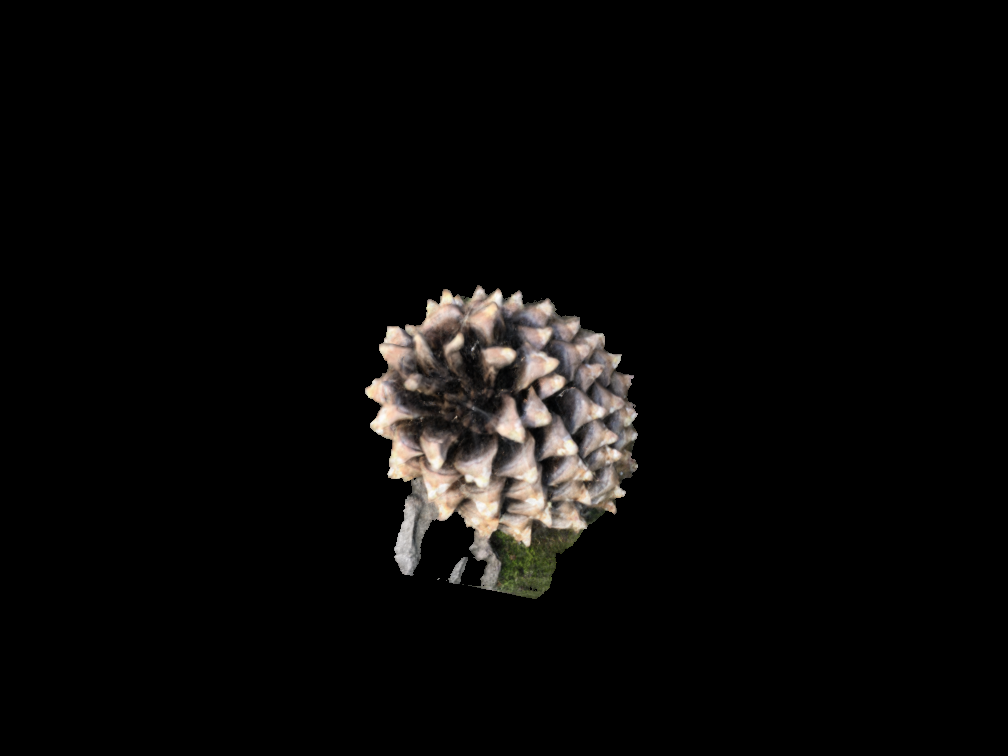}}
\subfloat{\includegraphics[width=0.11\linewidth]{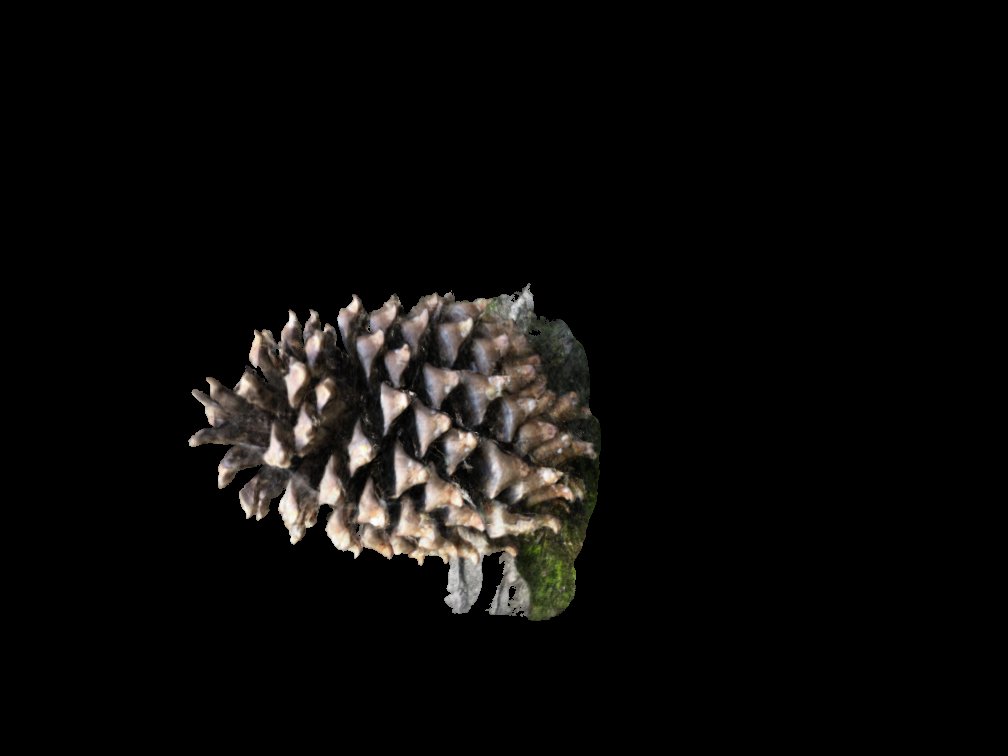}}
\subfloat{\includegraphics[width=0.11\linewidth]{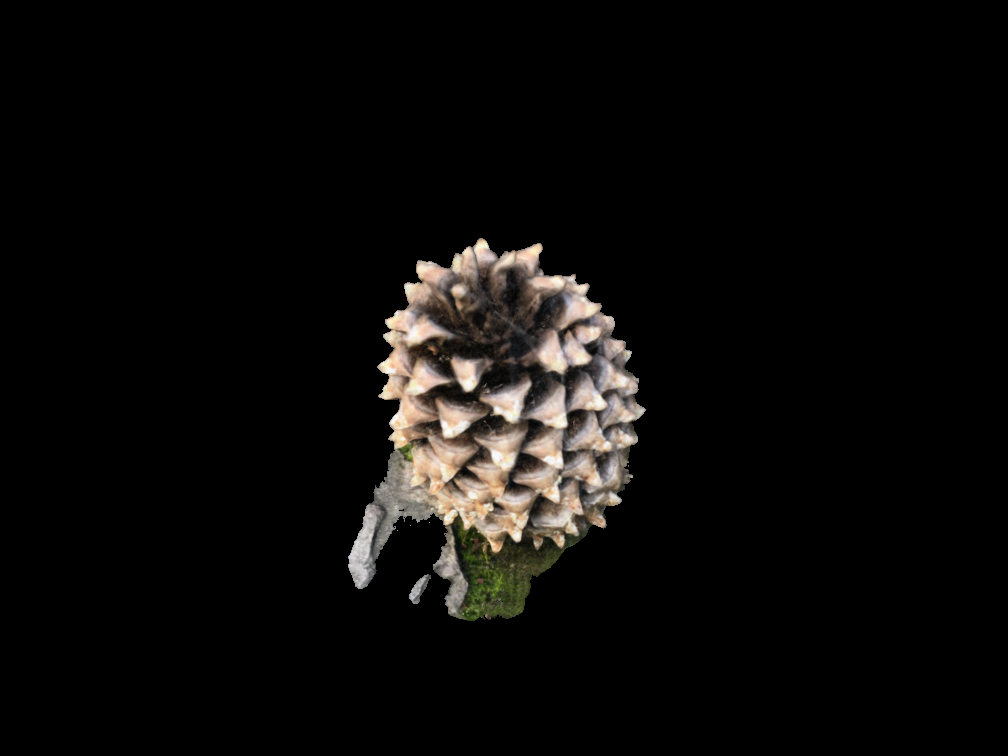}}
\subfloat{\includegraphics[width=0.11\linewidth]{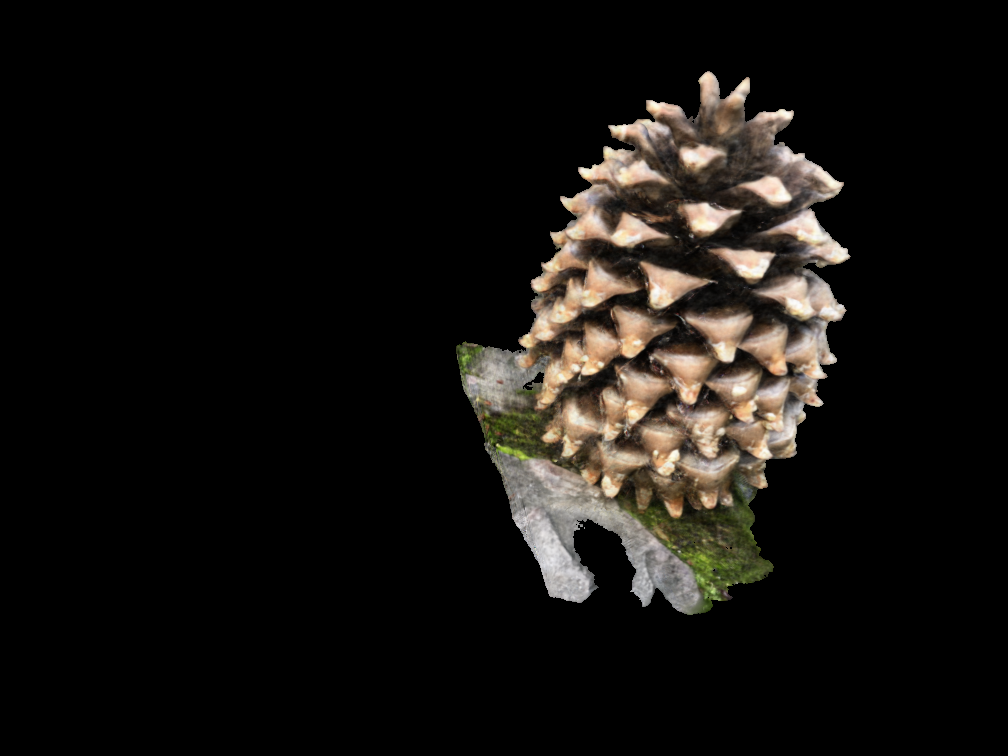}}
\subfloat{\includegraphics[width=0.11\linewidth]{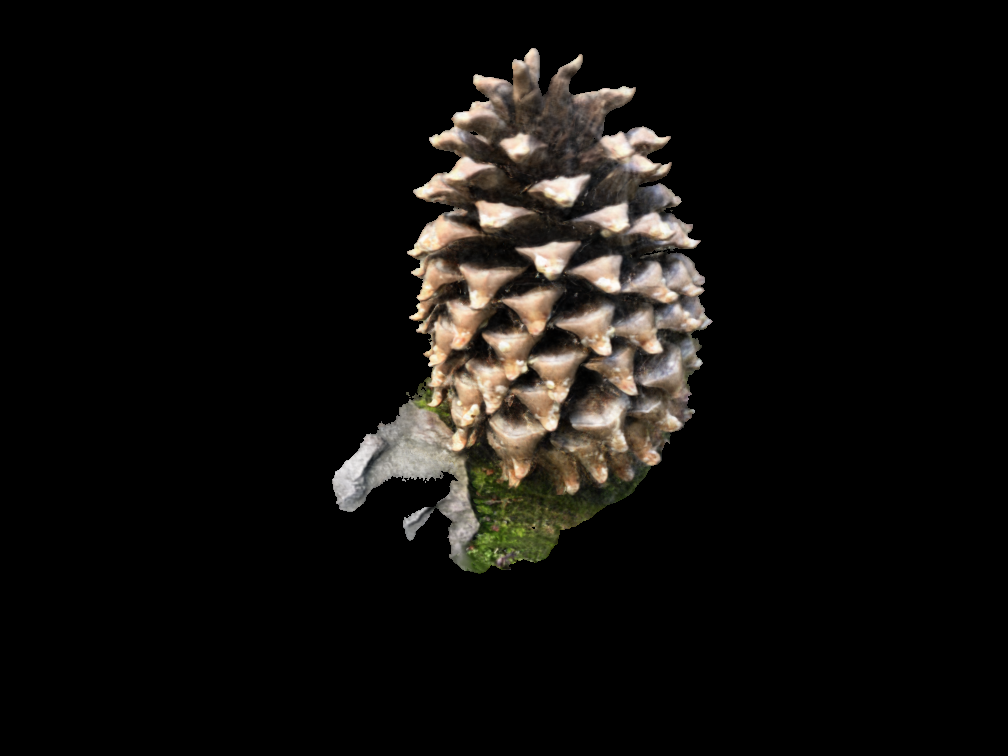}}
\subfloat{\includegraphics[width=0.11\linewidth]{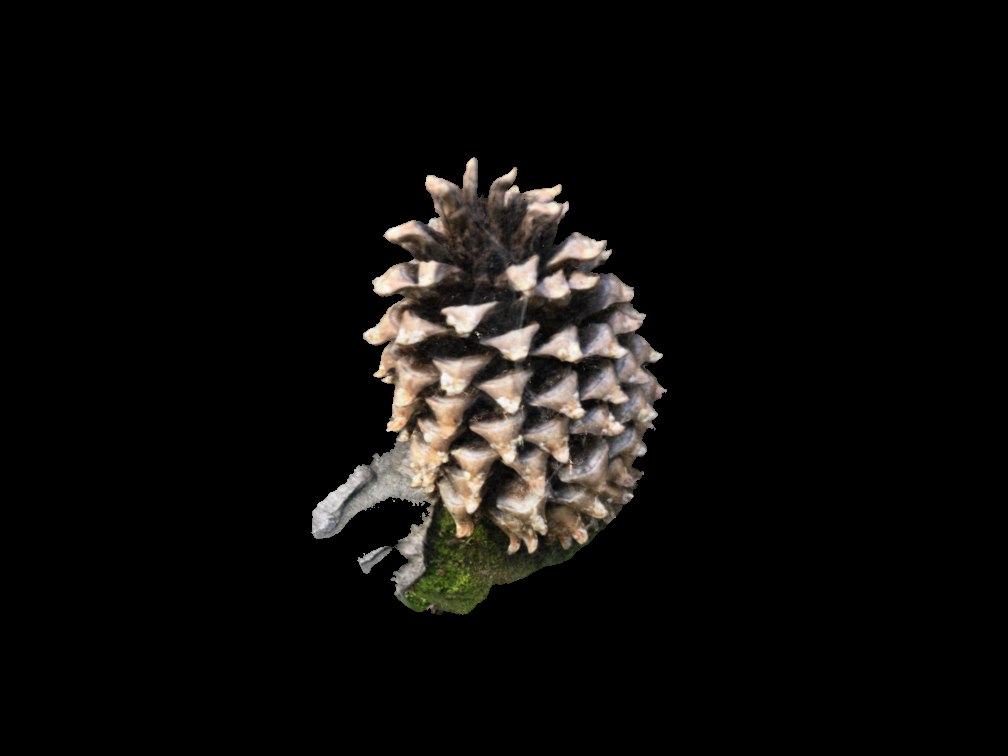}}
\subfloat{\includegraphics[width=0.11\linewidth]{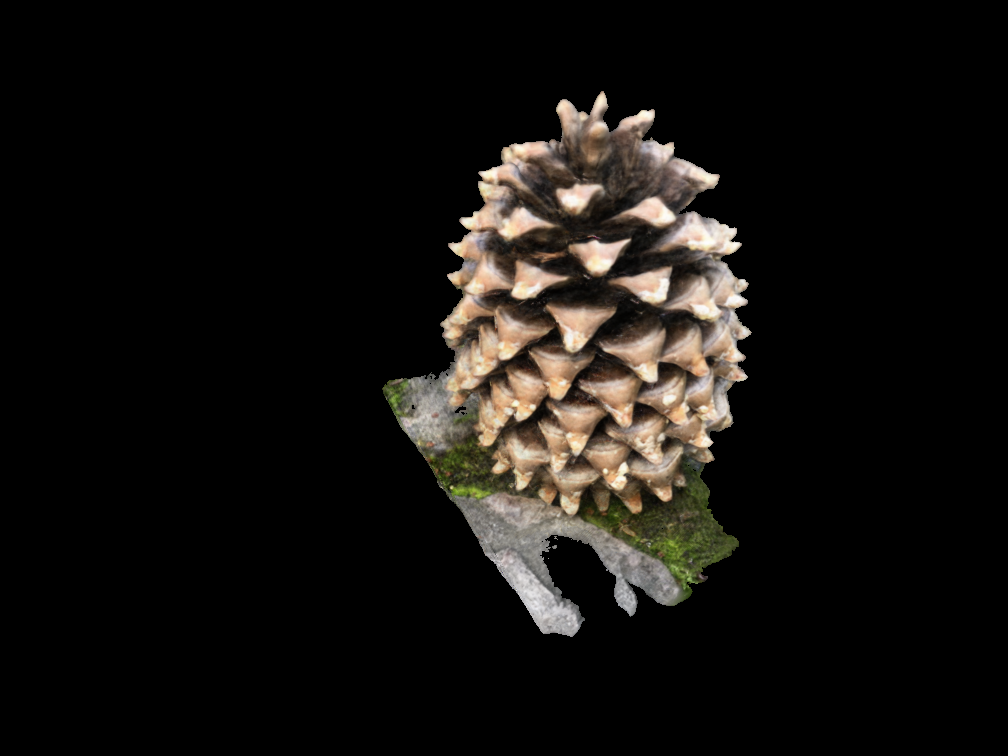}}
\end{minipage}

\begin{minipage}{0.2\linewidth}
\begin{tabular}{cc}
\multicolumn{2}{c}{\textbf{NERF}}                                                \\ \hline
PSNR & Time \\
28.44 & 23h51m
\end{tabular}
\end{minipage}
\begin{minipage}{0.8\linewidth}
\subfloat{\includegraphics[width=0.11\linewidth]{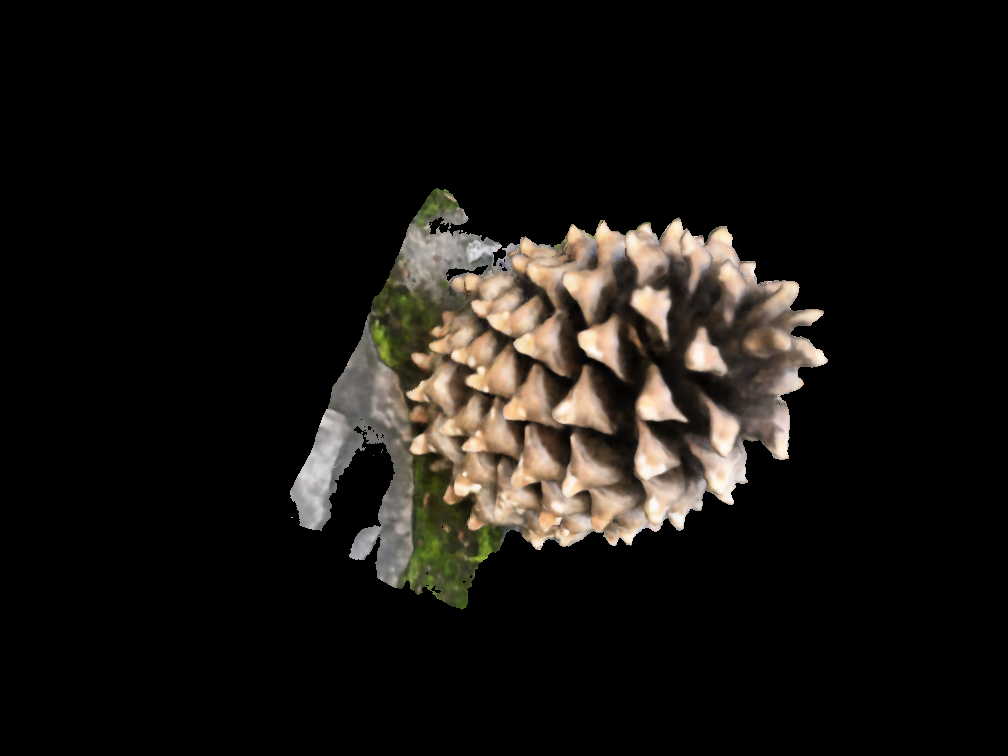}}
\subfloat{\includegraphics[width=0.11\linewidth]{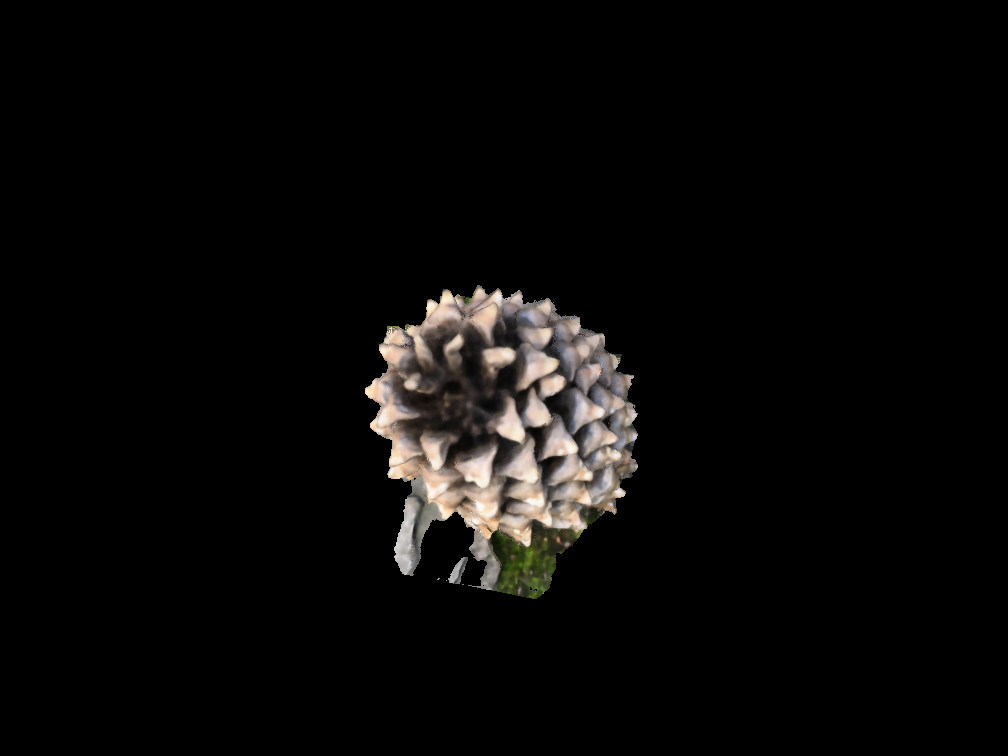}}
\subfloat{\includegraphics[width=0.11\linewidth]{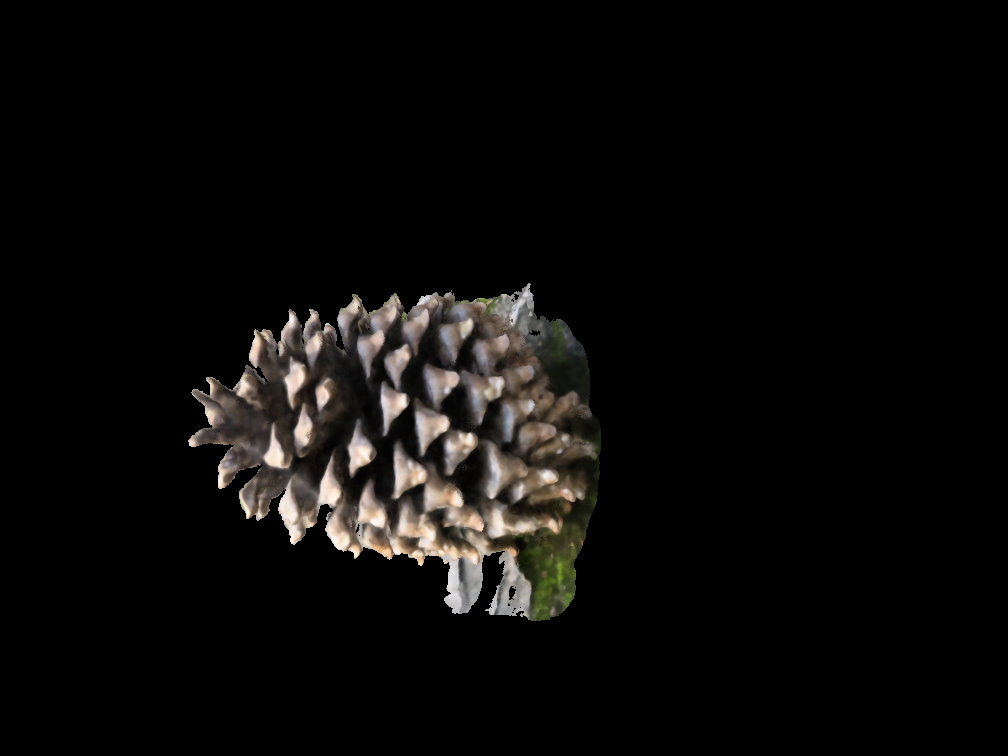}}
\subfloat{\includegraphics[width=0.11\linewidth]{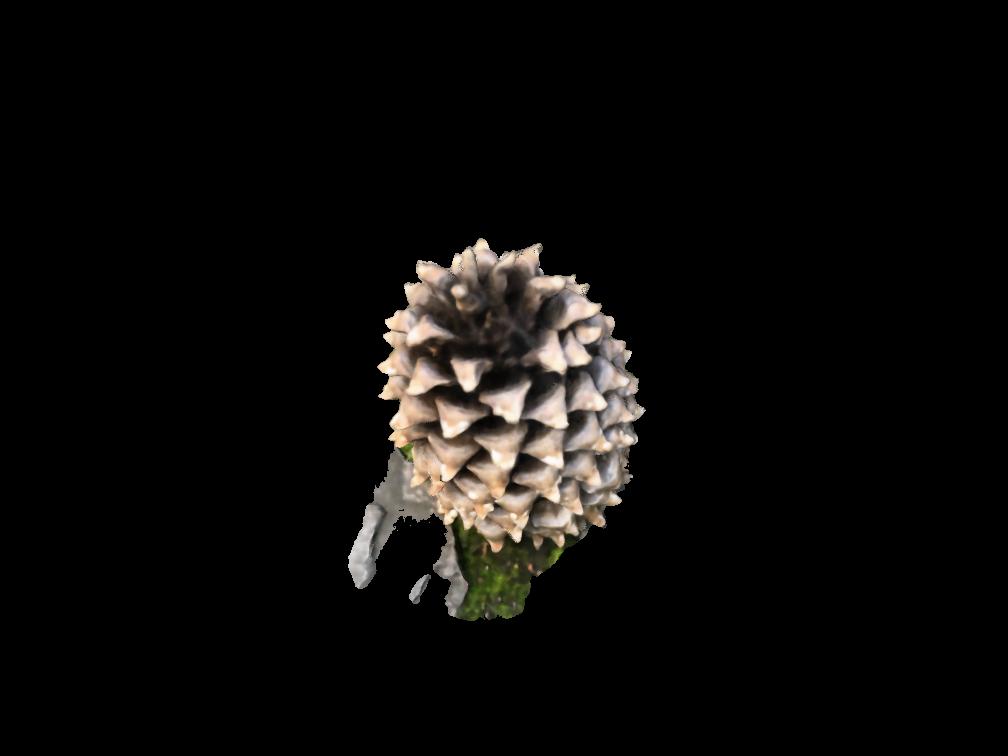}}
\subfloat{\includegraphics[width=0.11\linewidth]{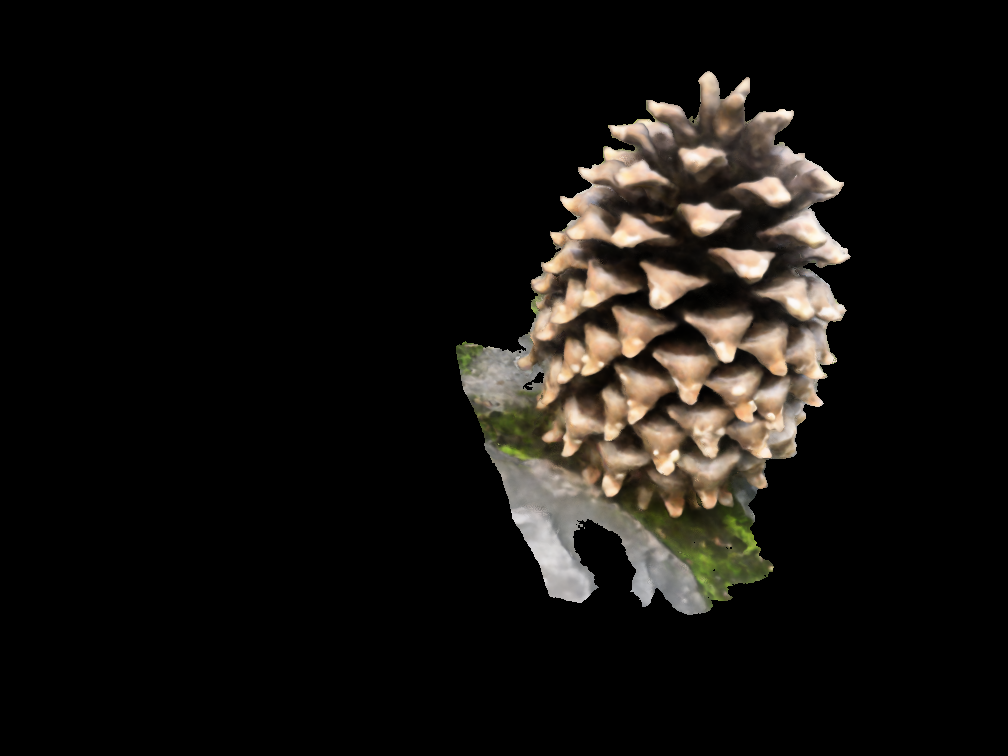}}
\subfloat{\includegraphics[width=0.11\linewidth]{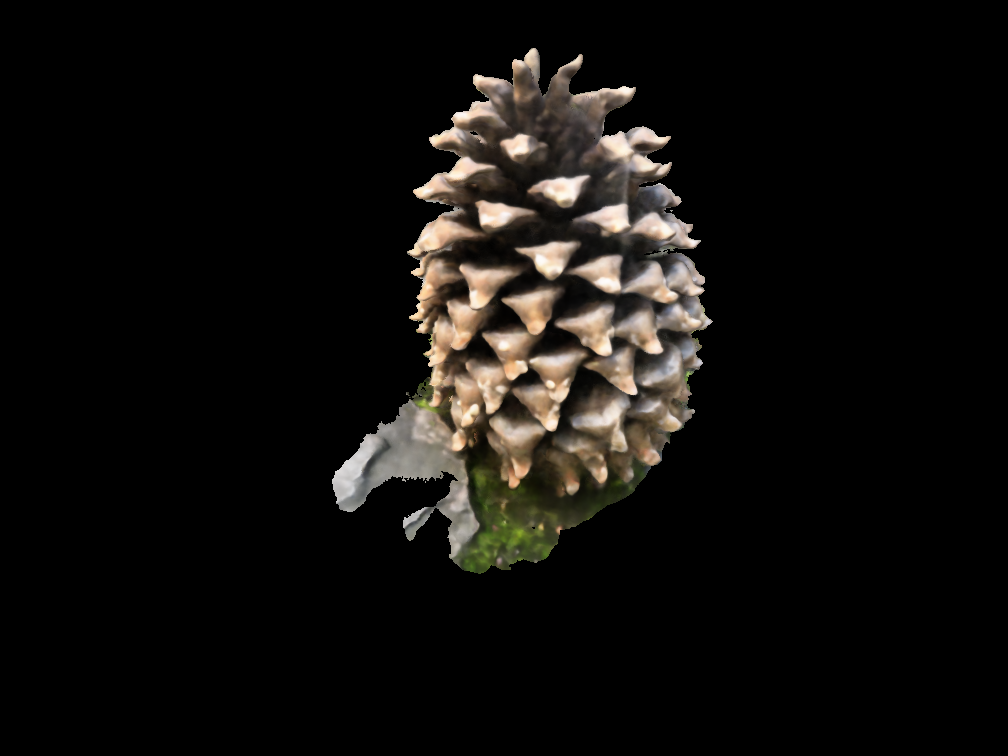}}
\subfloat{\includegraphics[width=0.11\linewidth]{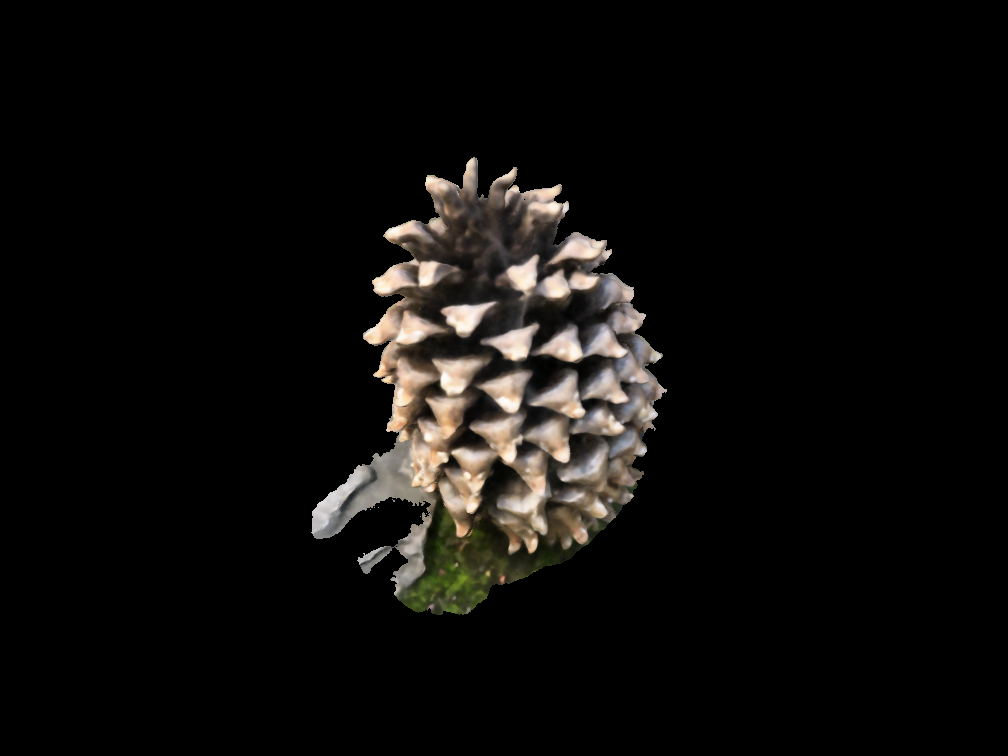}}
\subfloat{\includegraphics[width=0.11\linewidth]{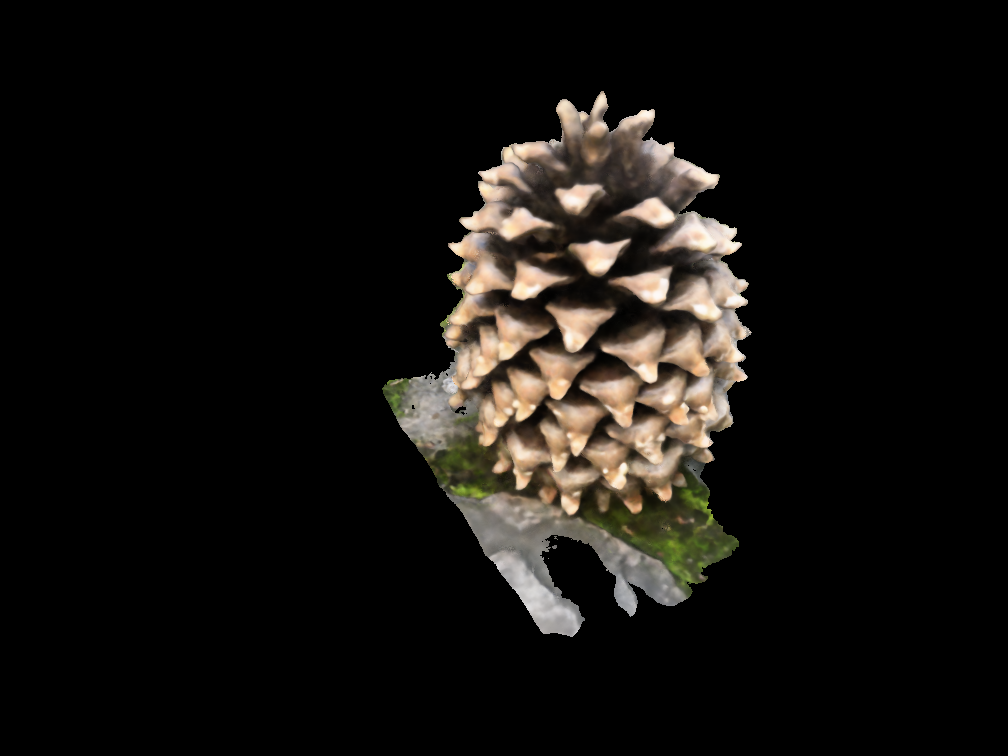}}
\end{minipage}
\begin{minipage}{0.2\linewidth}
\begin{tabular}{cc}
\multicolumn{2}{c}{\textbf{Ours}}                                                \\ \hline
PSNR & Time \\
25.51 & 28m59s
\end{tabular}
\end{minipage}
\begin{minipage}{0.8\linewidth}
\subfloat{\includegraphics[width=0.11\linewidth]{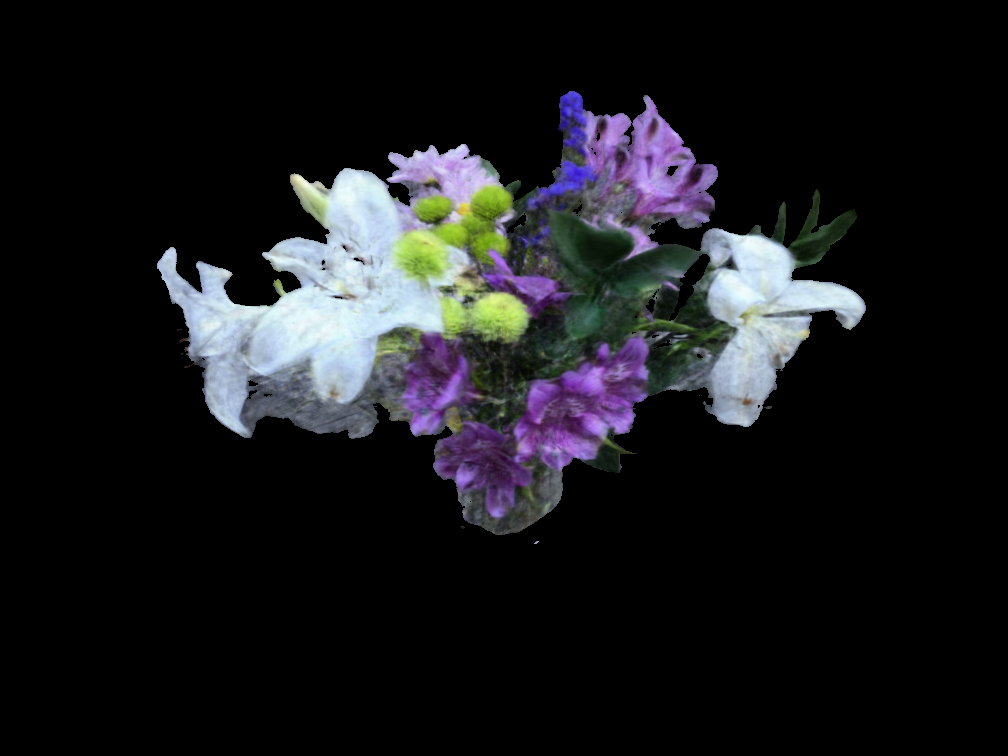}}
\subfloat{\includegraphics[width=0.11\linewidth]{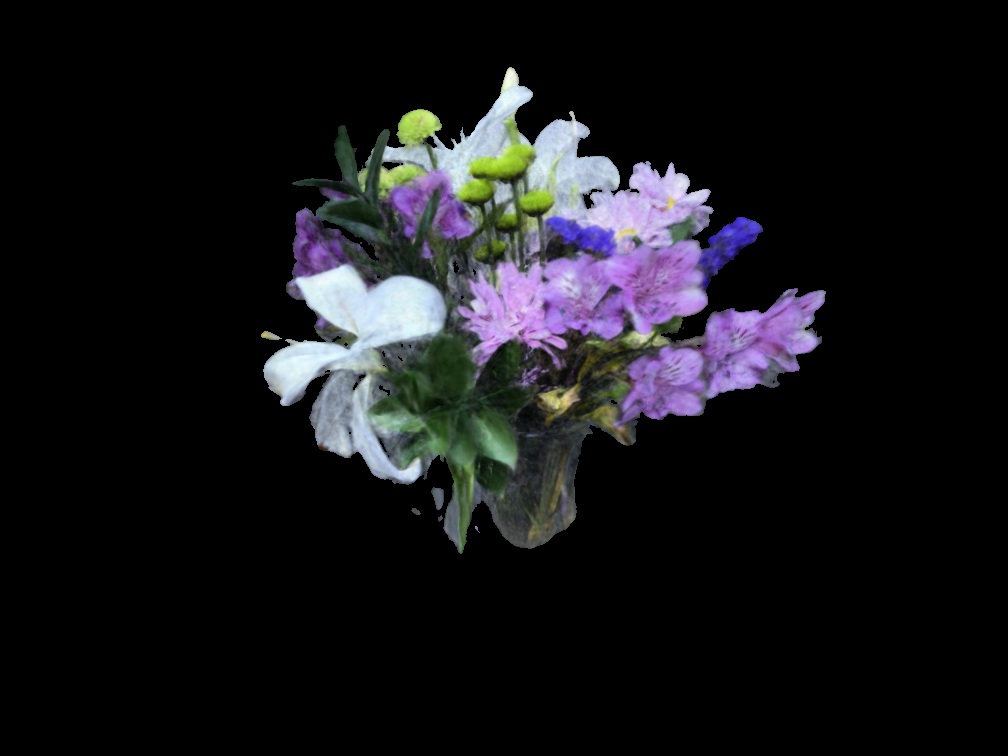}}
\subfloat{\includegraphics[width=0.11\linewidth]{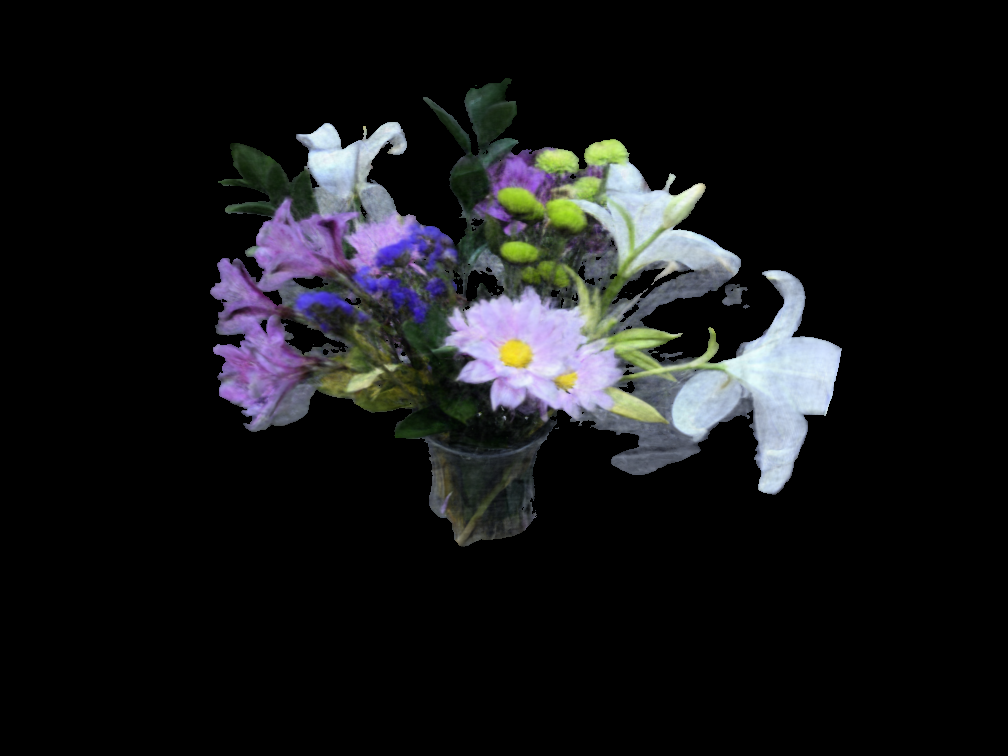}}
\subfloat{\includegraphics[width=0.11\linewidth]{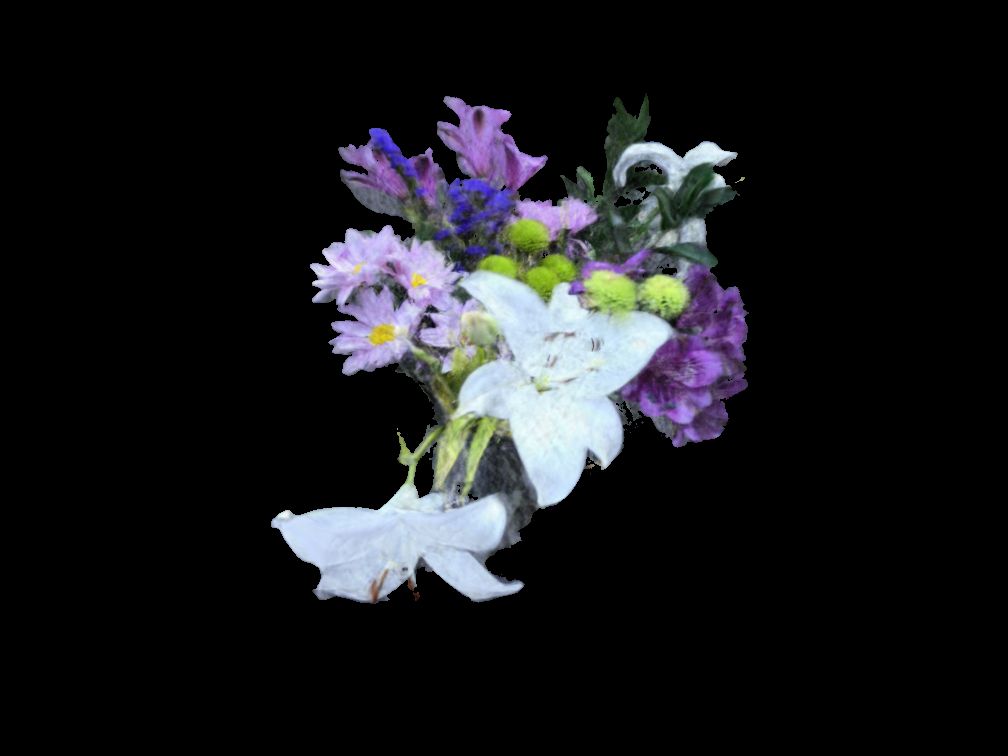}}
\subfloat{\includegraphics[width=0.11\linewidth]{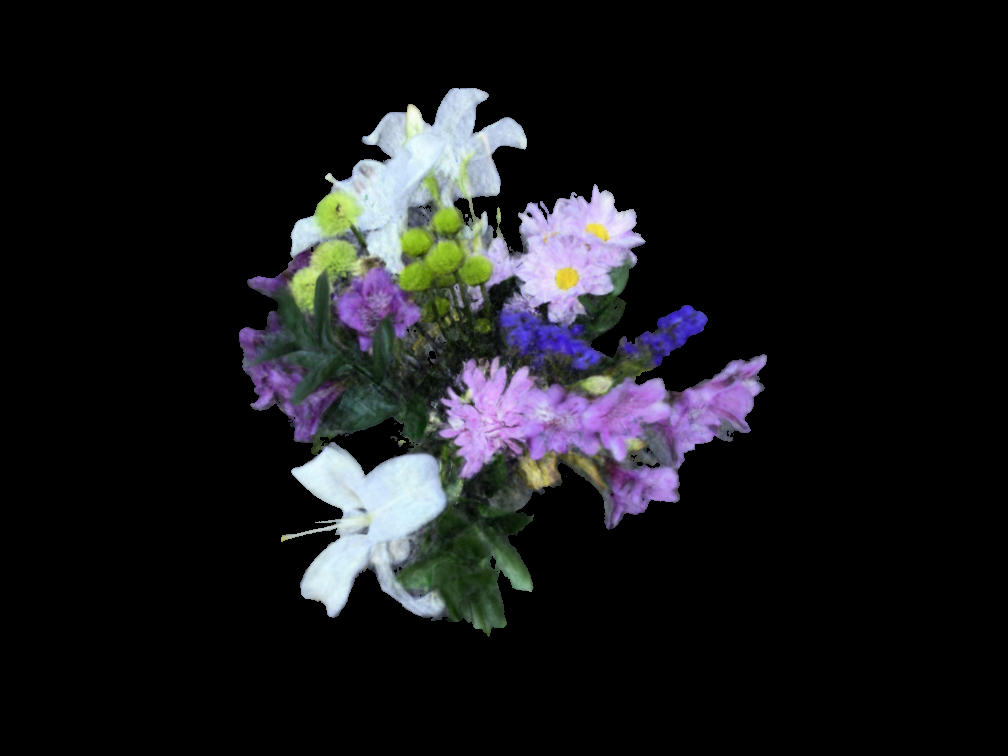}}
\subfloat{\includegraphics[width=0.11\linewidth]{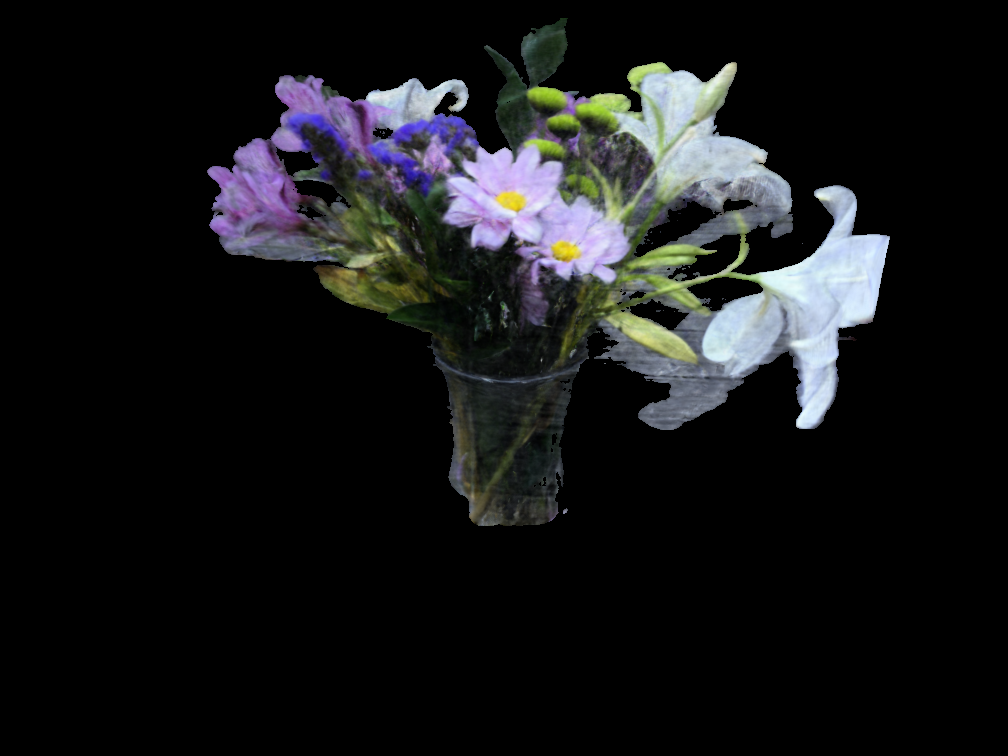}}
\subfloat{\includegraphics[width=0.11\linewidth]{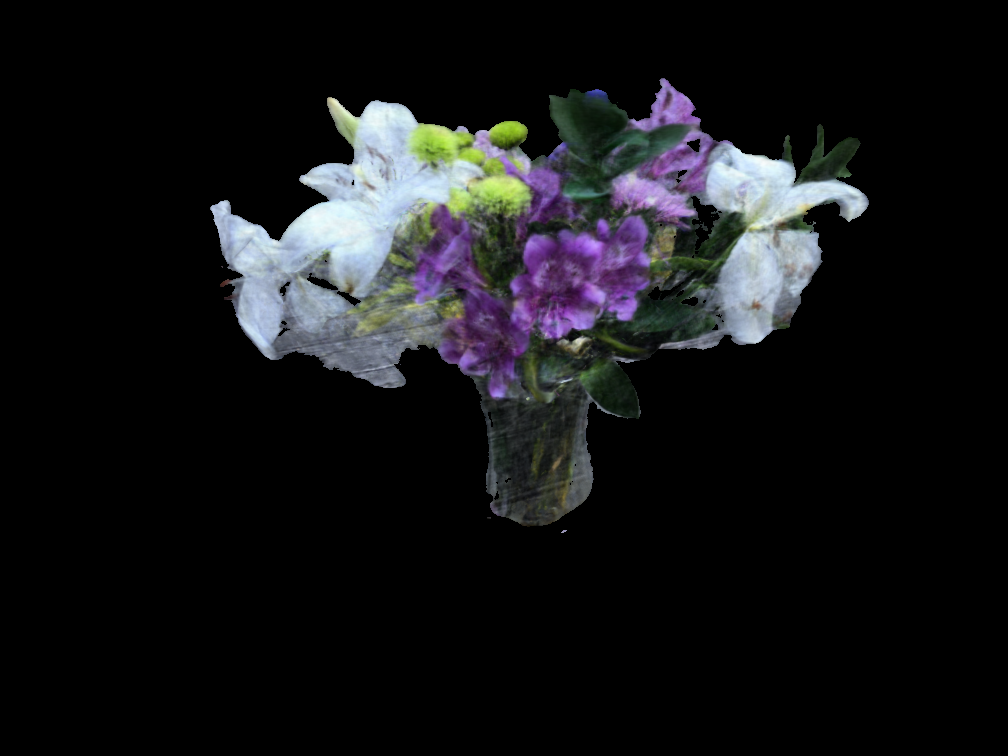}}
\subfloat{\includegraphics[width=0.11\linewidth]{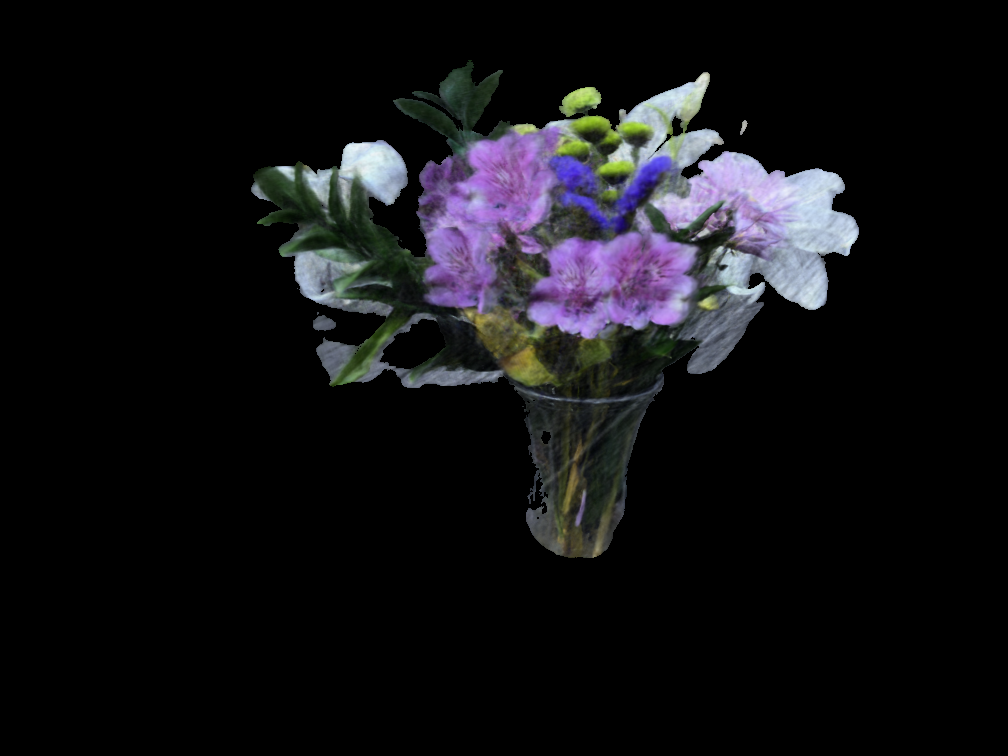}}
\end{minipage}

\begin{minipage}{0.2\linewidth}
\begin{tabular}{cc}
\multicolumn{2}{c}{\textbf{NERF}}                                                \\ \hline
PSNR & Time \\
26.10 & 20h15m
\end{tabular}
\end{minipage}
\begin{minipage}{0.8\linewidth}
\subfloat{\includegraphics[width=0.11\linewidth]{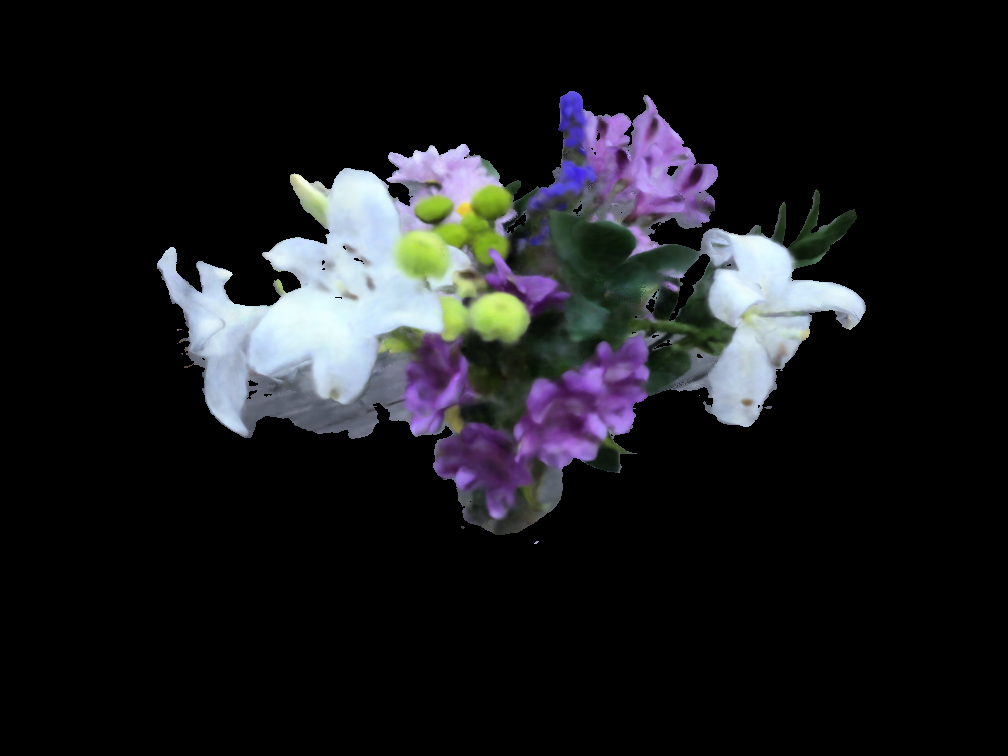}}
\subfloat{\includegraphics[width=0.11\linewidth]{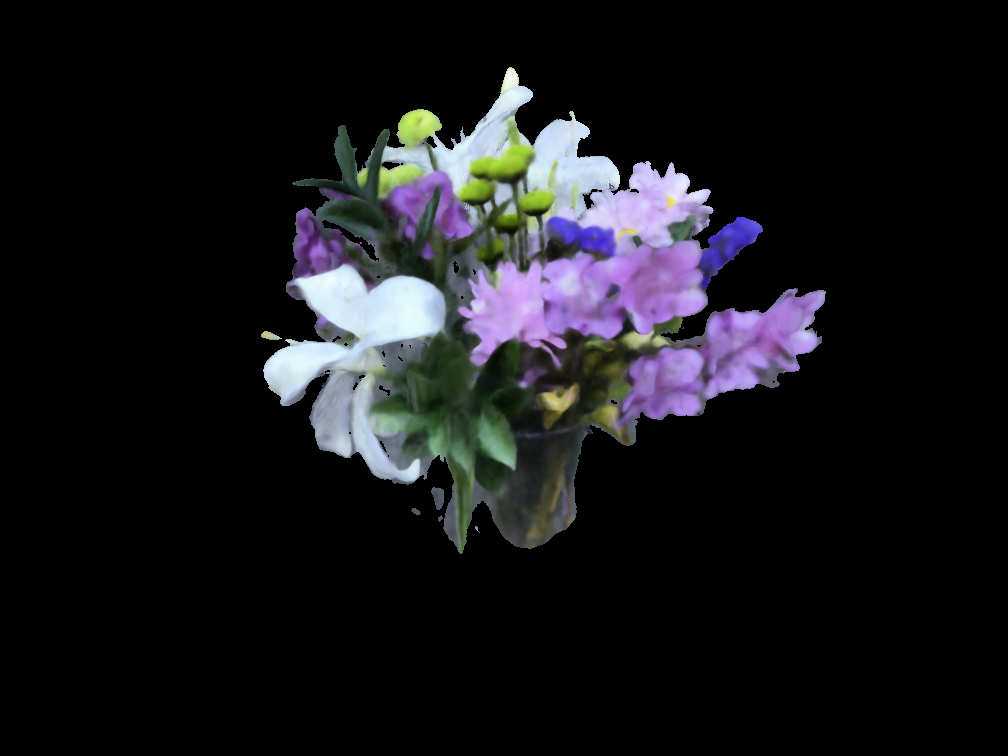}}
\subfloat{\includegraphics[width=0.11\linewidth]{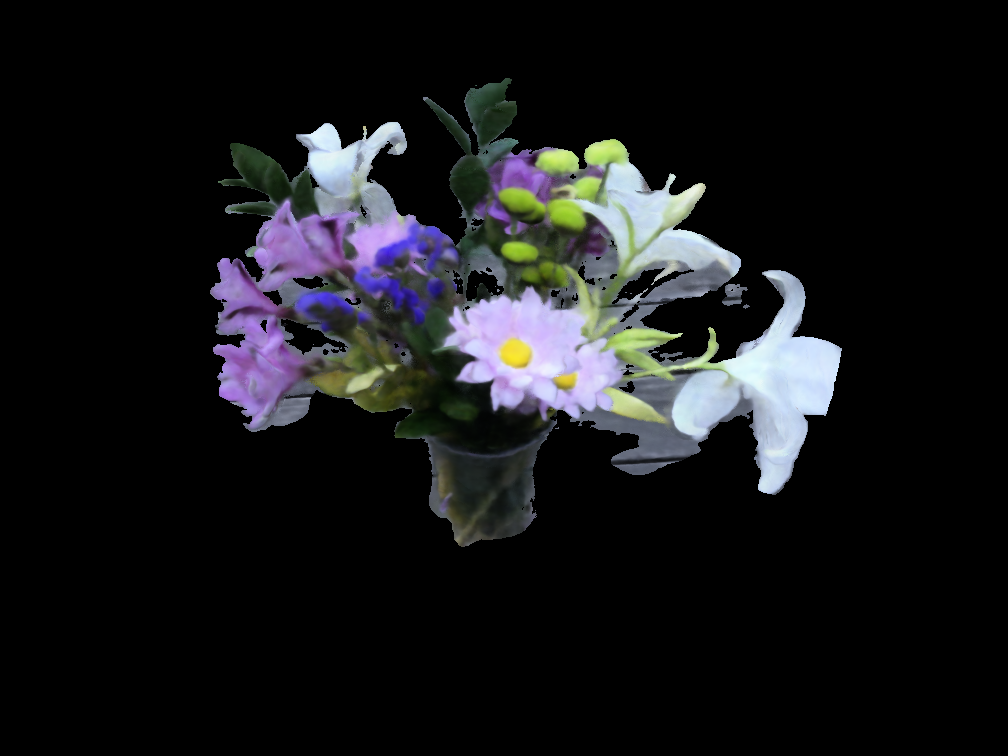}}
\subfloat{\includegraphics[width=0.11\linewidth]{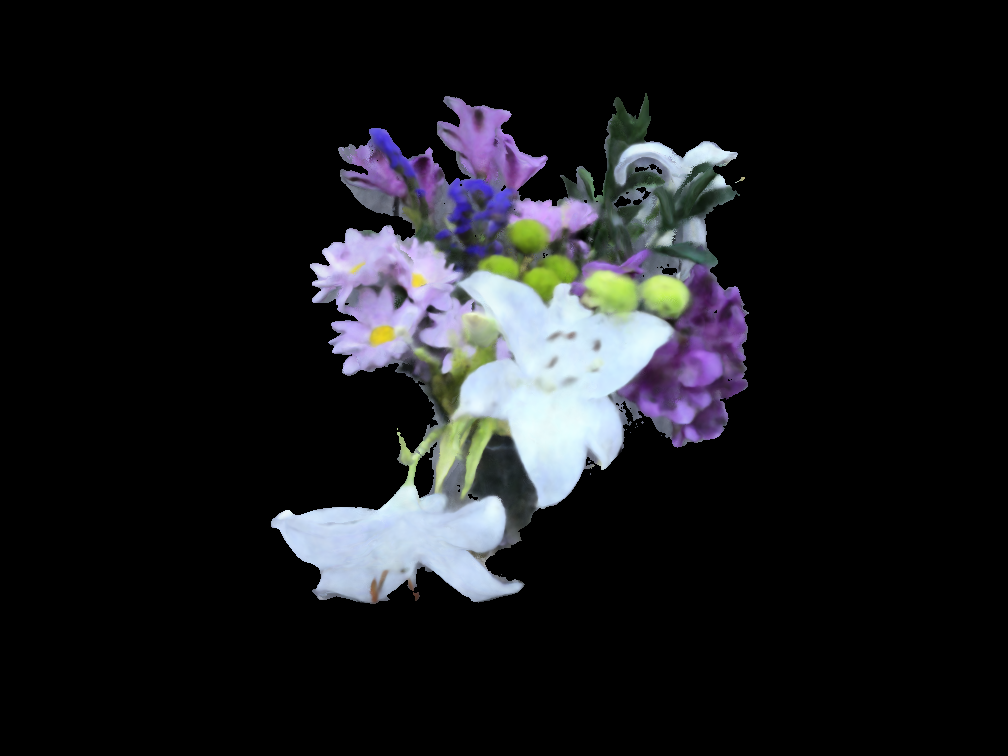}}
\subfloat{\includegraphics[width=0.11\linewidth]{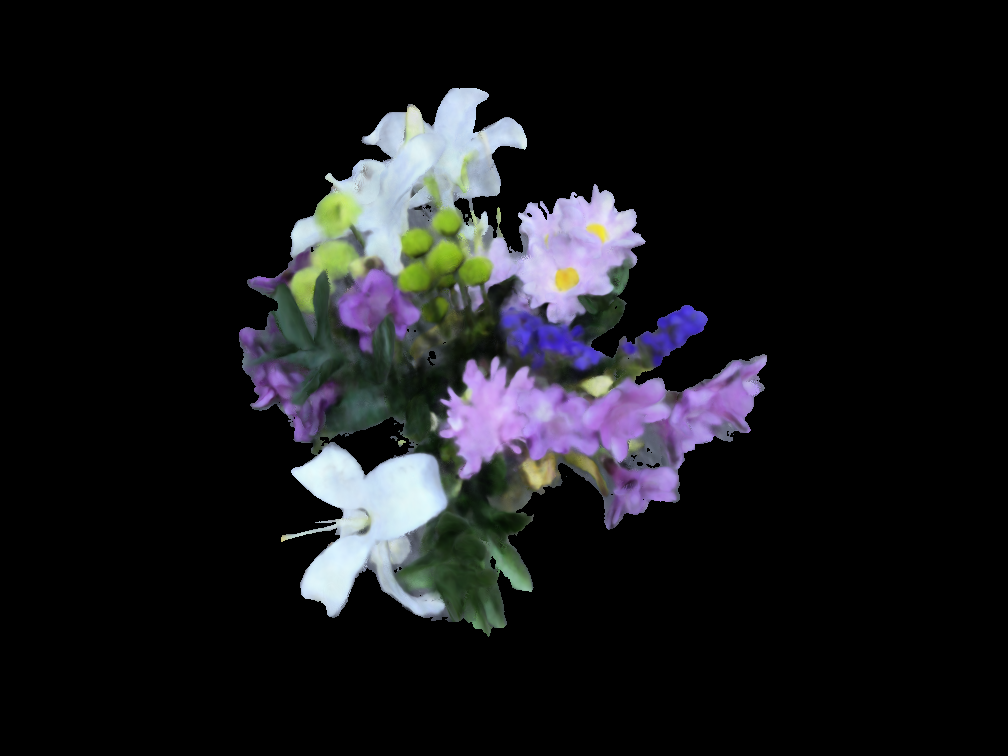}}
\subfloat{\includegraphics[width=0.11\linewidth]{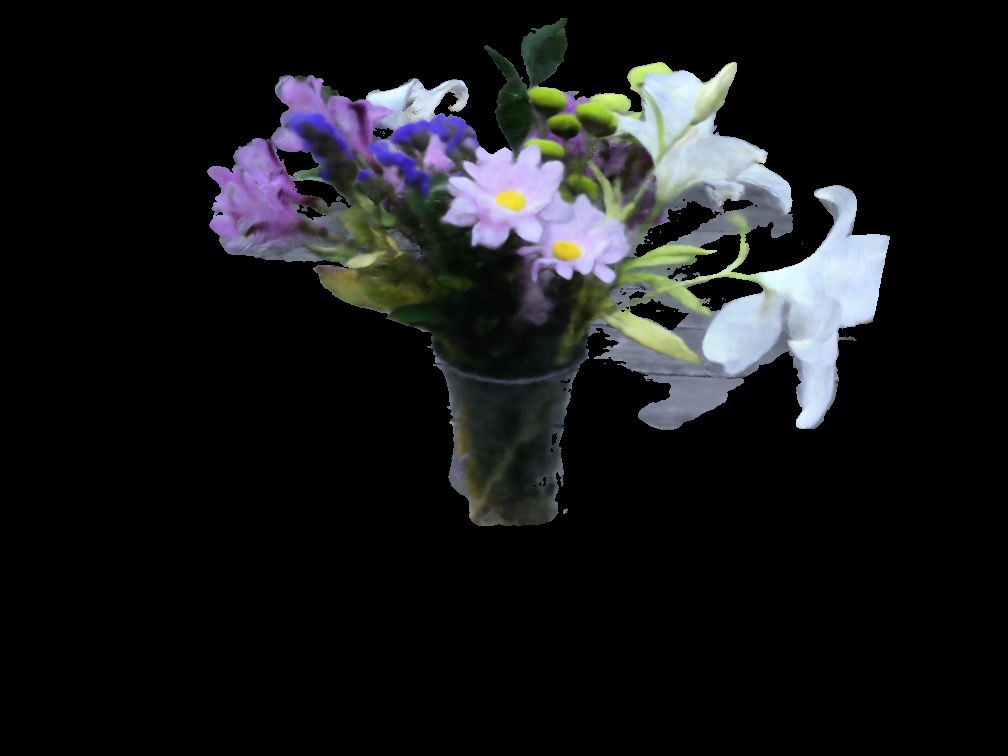}}
\subfloat{\includegraphics[width=0.11\linewidth]{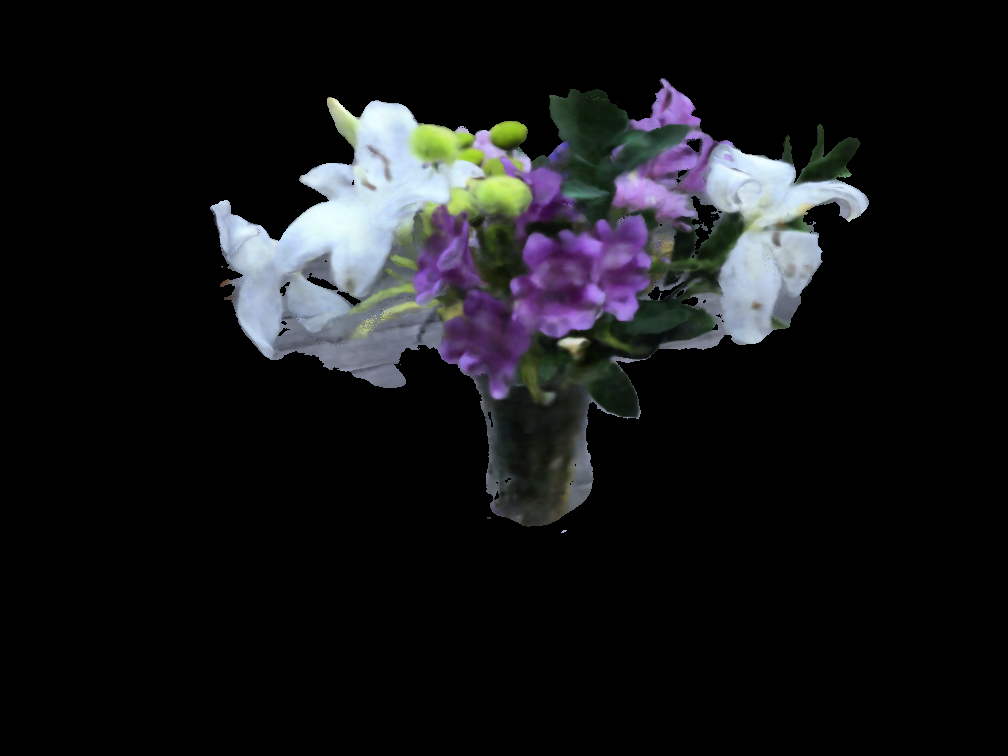}}
\subfloat{\includegraphics[width=0.11\linewidth]{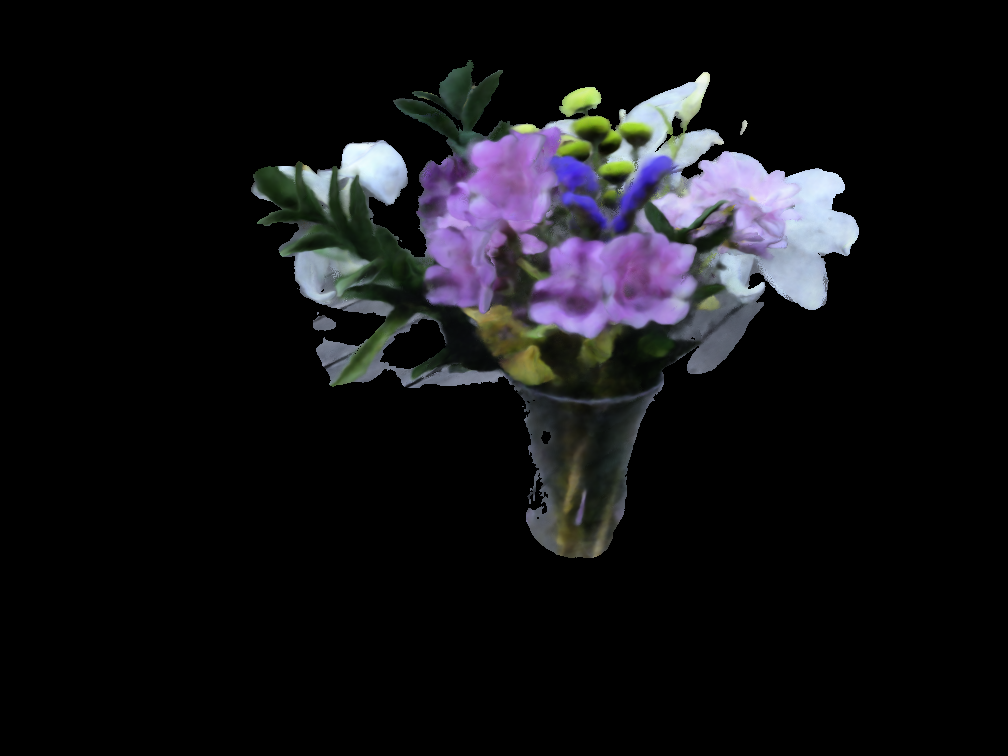}}
\end{minipage}

\begin{minipage}{0.2\linewidth}
\begin{tabular}{cc}
\multicolumn{2}{c}{\textbf{Ours}}                                                \\ \hline
PSNR & Time \\
28.25 & 4m26s
\end{tabular}
\end{minipage}
\begin{minipage}{0.8\linewidth}
\subfloat{\includegraphics[width=0.11\linewidth]{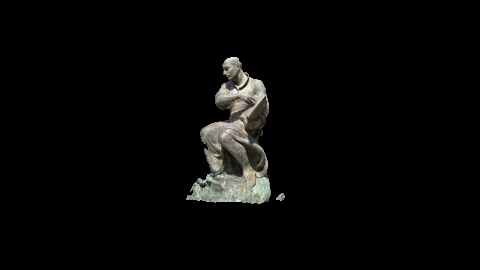}}
\subfloat{\includegraphics[width=0.11\linewidth]{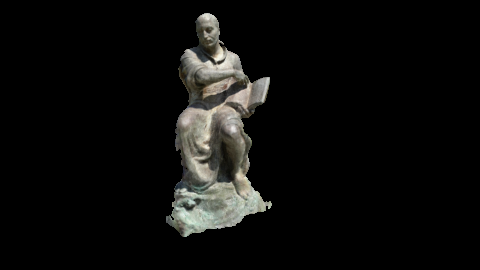}}
\subfloat{\includegraphics[width=0.11\linewidth]{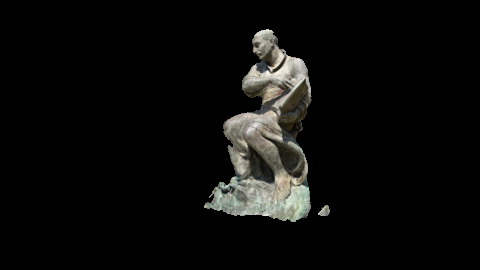}}
\subfloat{\includegraphics[width=0.11\linewidth]{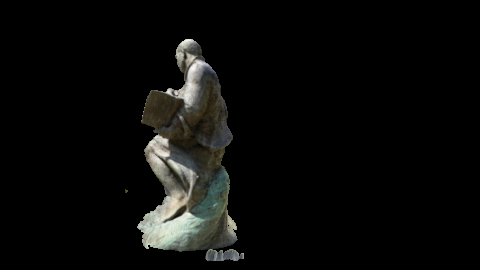}}
\subfloat{\includegraphics[width=0.11\linewidth]{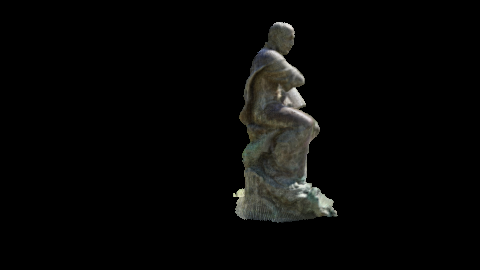}}
\subfloat{\includegraphics[width=0.11\linewidth]{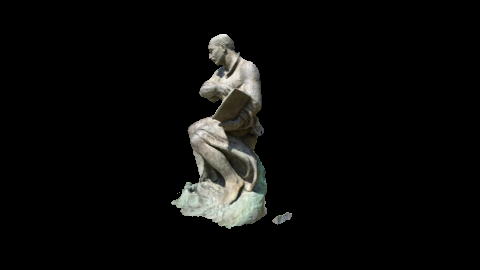}}
\subfloat{\includegraphics[width=0.11\linewidth]{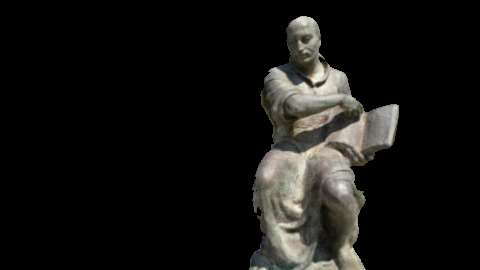}}
\subfloat{\includegraphics[width=0.11\linewidth]{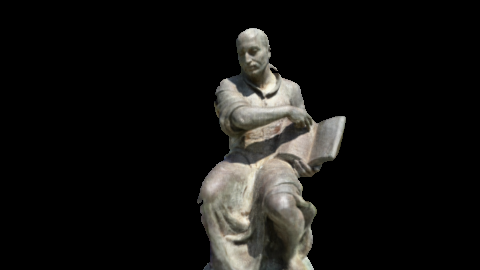}}
\end{minipage}

\begin{minipage}{0.2\linewidth}
\begin{tabular}{cc}
\multicolumn{2}{c}{\textbf{NERF}}                                                \\ \hline
PSNR & Time \\
27.95 & 24h17m
\end{tabular}
\end{minipage}
\begin{minipage}{0.8\linewidth}
\subfloat{\includegraphics[width=0.11\linewidth]{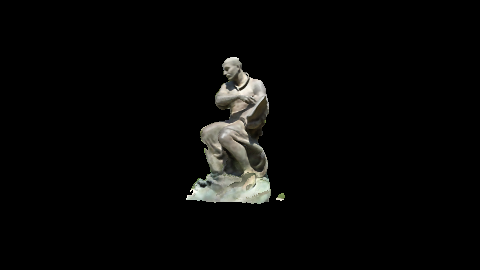}}
\subfloat{\includegraphics[width=0.11\linewidth]{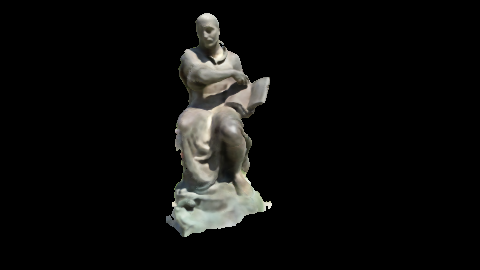}}
\subfloat{\includegraphics[width=0.11\linewidth]{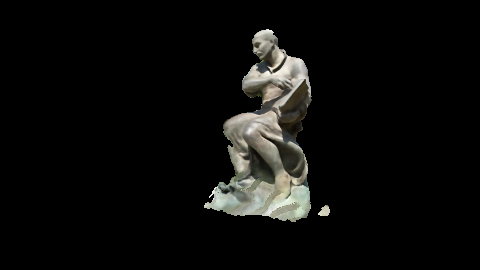}}
\subfloat{\includegraphics[width=0.11\linewidth]{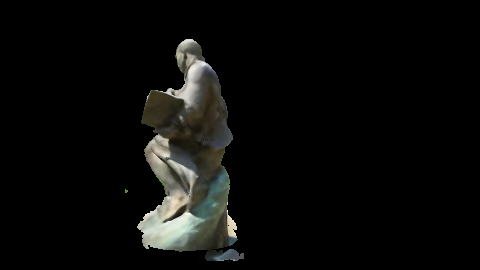}}
\subfloat{\includegraphics[width=0.11\linewidth]{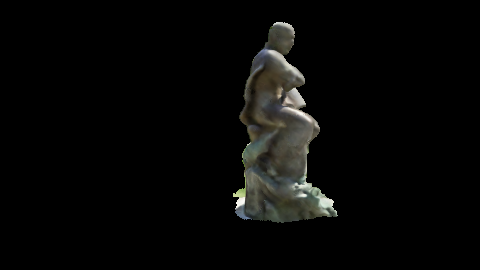}}
\subfloat{\includegraphics[width=0.11\linewidth]{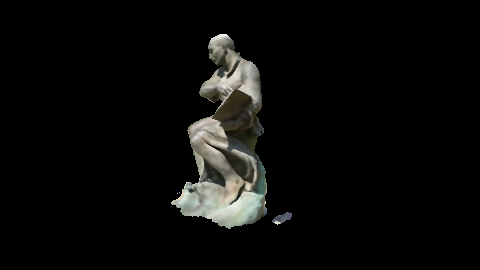}}
\subfloat{\includegraphics[width=0.11\linewidth]{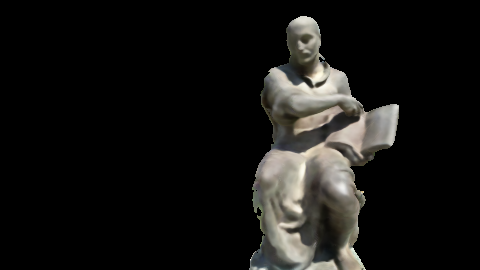}}
\subfloat{\includegraphics[width=0.11\linewidth]{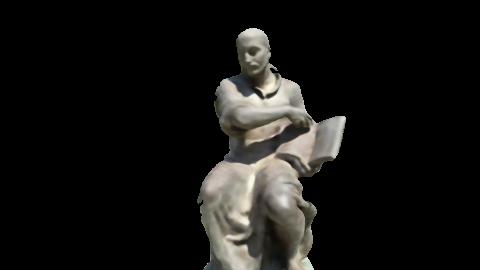}}
\end{minipage}

\begin{minipage}{0.2\linewidth}
\begin{tabular}{cc}
\multicolumn{2}{c}{\textbf{Ours}}                                                \\ \hline
PSNR & Time \\
31.82 & 3m23s
\end{tabular}
\end{minipage}
\begin{minipage}{0.8\linewidth}
\subfloat{\includegraphics[width=0.11\linewidth]{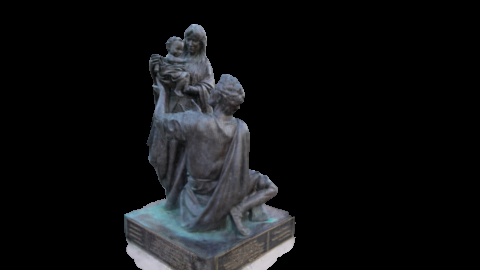}}
\subfloat{\includegraphics[width=0.11\linewidth]{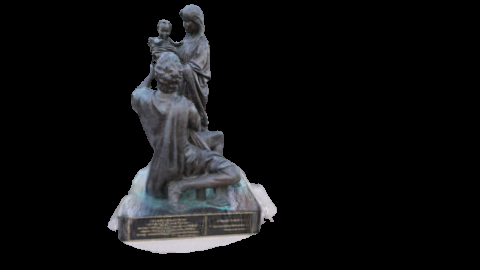}}
\subfloat{\includegraphics[width=0.11\linewidth]{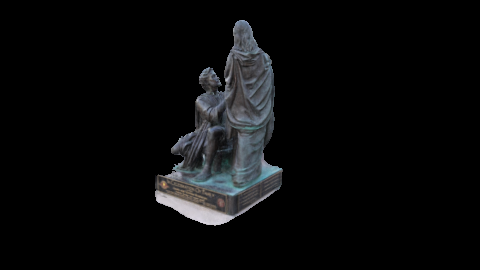}}
\subfloat{\includegraphics[width=0.11\linewidth]{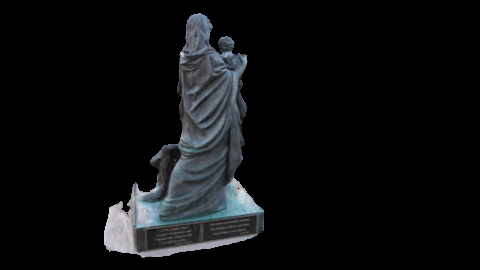}}
\subfloat{\includegraphics[width=0.11\linewidth]{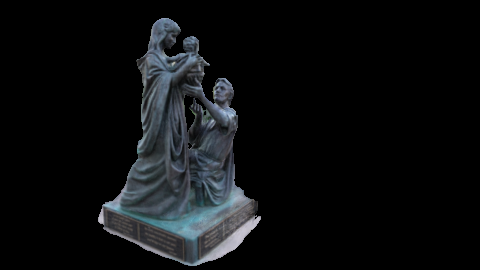}}
\subfloat{\includegraphics[width=0.11\linewidth]{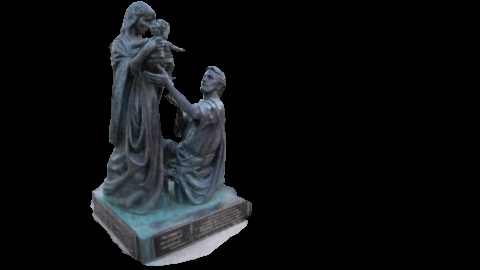}}
\subfloat{\includegraphics[width=0.11\linewidth]{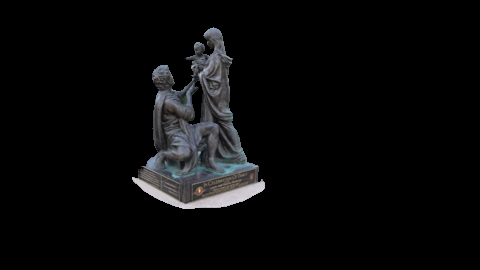}}
\subfloat{\includegraphics[width=0.11\linewidth]{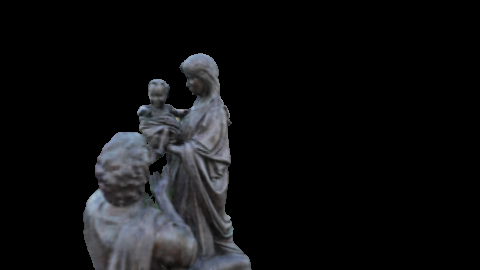}}
\end{minipage}

\begin{minipage}{0.2\linewidth}
\begin{tabular}{cc}
\multicolumn{2}{c}{\textbf{NERF}}                                                \\ \hline
PSNR & Time \\
30.49 & 20h18m
\end{tabular}
\end{minipage}
\begin{minipage}{0.8\linewidth}
\subfloat{\includegraphics[width=0.11\linewidth]{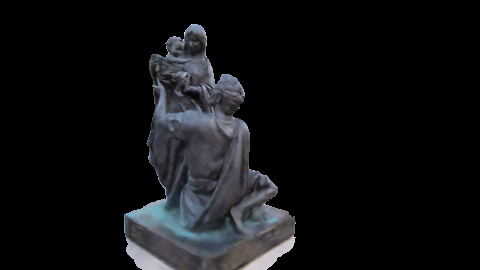}}
\subfloat{\includegraphics[width=0.11\linewidth]{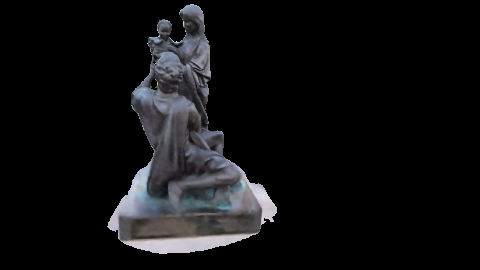}}
\subfloat{\includegraphics[width=0.11\linewidth]{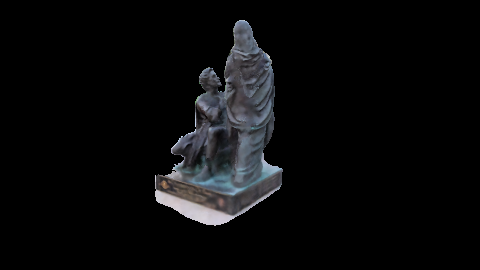}}
\subfloat{\includegraphics[width=0.11\linewidth]{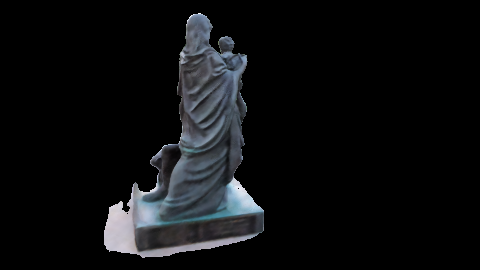}}
\subfloat{\includegraphics[width=0.11\linewidth]{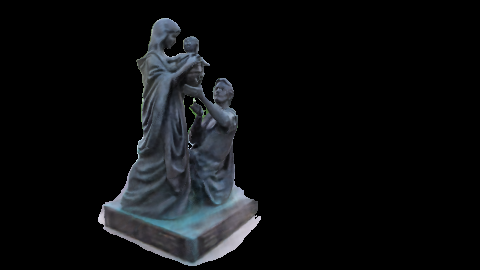}}
\subfloat{\includegraphics[width=0.11\linewidth]{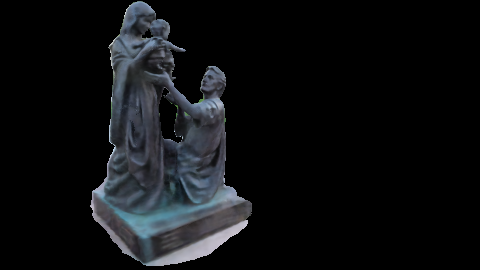}}
\subfloat{\includegraphics[width=0.11\linewidth]{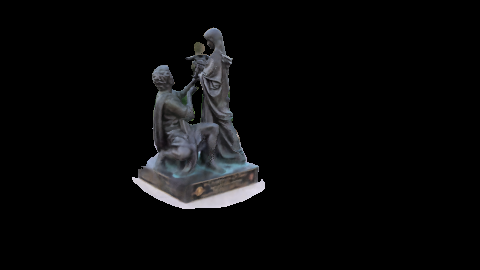}}
\subfloat{\includegraphics[width=0.11\linewidth]{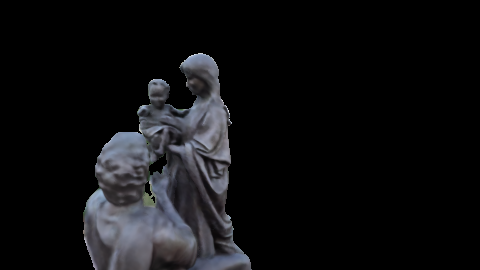}}
\end{minipage}

\begin{minipage}{0.2\linewidth}
\begin{tabular}{cc}
\multicolumn{2}{c}{\textbf{Ours}}                                                \\ \hline
PSNR & Time \\
23.84 & 34m53s
\end{tabular}
\end{minipage}
\begin{minipage}{0.8\linewidth}
\subfloat{\includegraphics[width=0.11\linewidth]{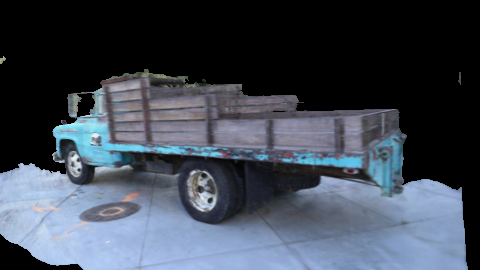}}
\subfloat{\includegraphics[width=0.11\linewidth]{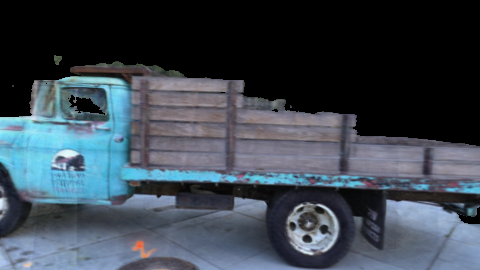}}
\subfloat{\includegraphics[width=0.11\linewidth]{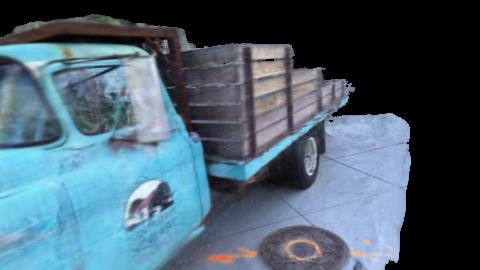}}
\subfloat{\includegraphics[width=0.11\linewidth]{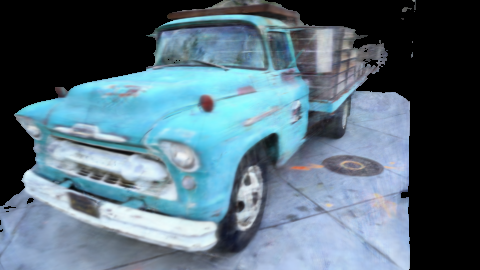}}
\subfloat{\includegraphics[width=0.11\linewidth]{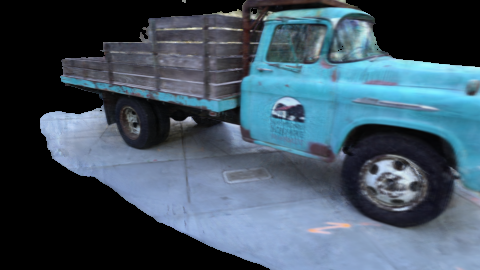}}
\subfloat{\includegraphics[width=0.11\linewidth]{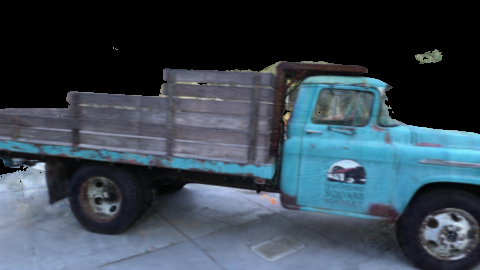}}
\subfloat{\includegraphics[width=0.11\linewidth]{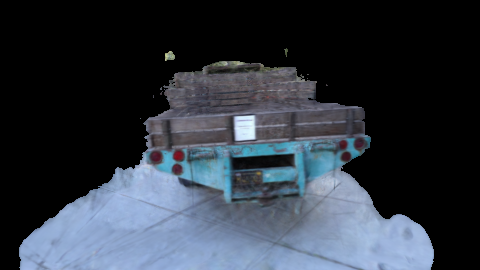}}
\subfloat{\includegraphics[width=0.11\linewidth]{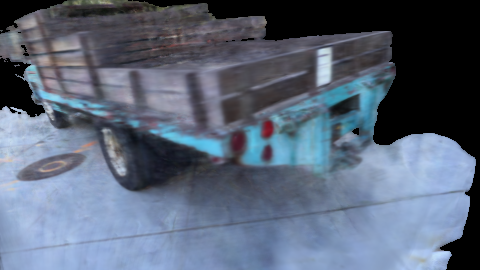}}
\end{minipage}

\begin{minipage}{0.2\linewidth}
\begin{tabular}{cc}
\multicolumn{2}{c}{\textbf{NERF}}                                                \\ \hline
PSNR & Time \\
23.51 & 18h16m
\end{tabular}
\end{minipage}
\begin{minipage}{0.8\linewidth}
\subfloat{\includegraphics[width=0.11\linewidth]{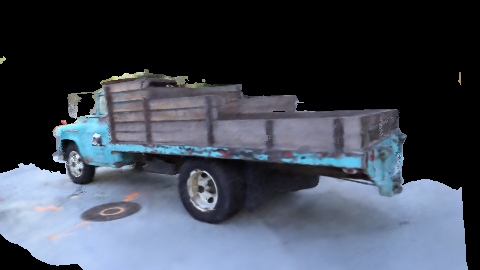}}
\subfloat{\includegraphics[width=0.11\linewidth]{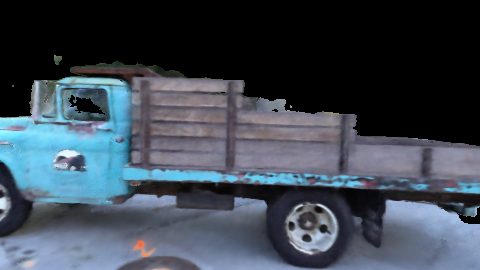}}
\subfloat{\includegraphics[width=0.11\linewidth]{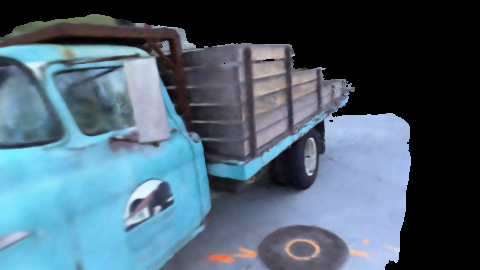}}
\subfloat{\includegraphics[width=0.11\linewidth]{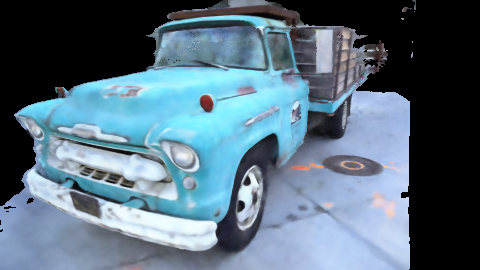}}
\subfloat{\includegraphics[width=0.11\linewidth]{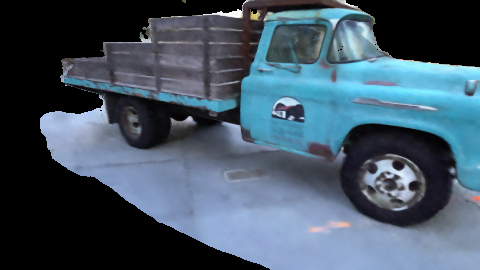}}
\subfloat{\includegraphics[width=0.11\linewidth]{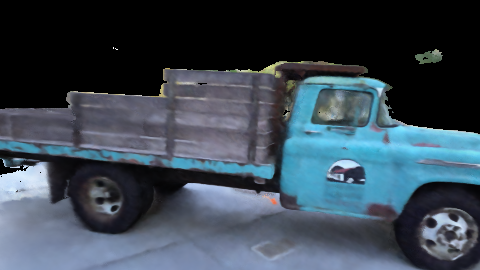}}
\subfloat{\includegraphics[width=0.11\linewidth]{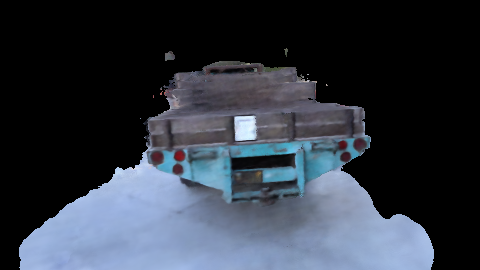}}
\subfloat{\includegraphics[width=0.11\linewidth]{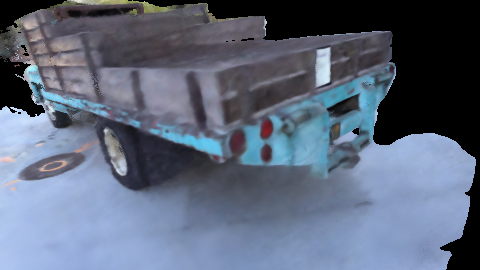}}
\end{minipage}

\begin{minipage}{0.2\linewidth}
\begin{tabular}{cc}
\multicolumn{2}{c}{\textbf{Ours}}                                                \\ \hline
PSNR & Time \\
23.45 & 12m32s
\end{tabular}
\end{minipage}
\begin{minipage}{0.8\linewidth}
\subfloat{\includegraphics[width=0.11\linewidth]{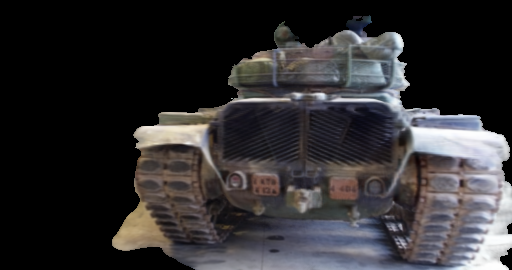}}
\subfloat{\includegraphics[width=0.11\linewidth]{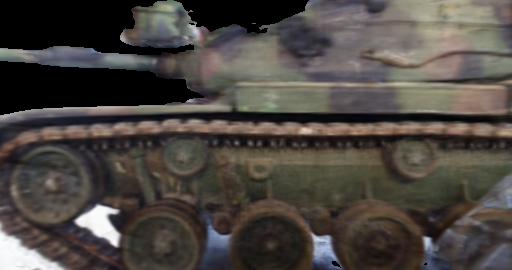}}
\subfloat{\includegraphics[width=0.11\linewidth]{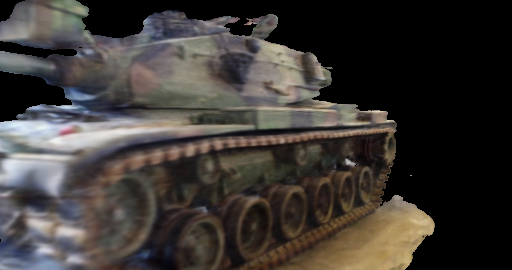}}
\subfloat{\includegraphics[width=0.11\linewidth]{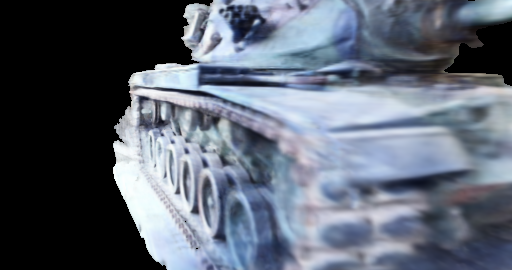}}
\subfloat{\includegraphics[width=0.11\linewidth]{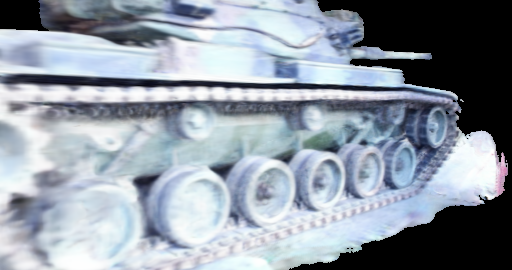}}
\subfloat{\includegraphics[width=0.11\linewidth]{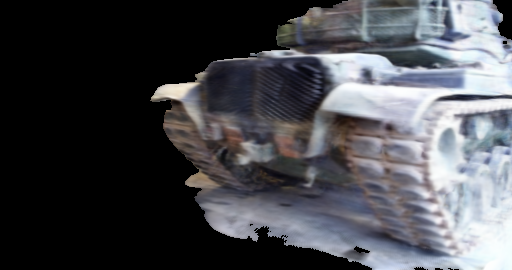}}
\subfloat{\includegraphics[width=0.11\linewidth]{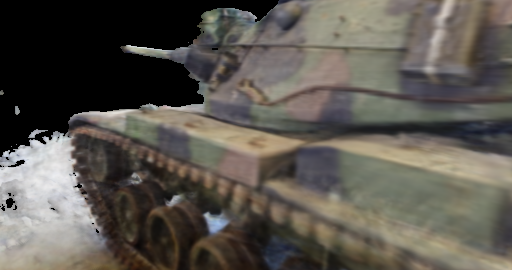}}
\subfloat{\includegraphics[width=0.11\linewidth]{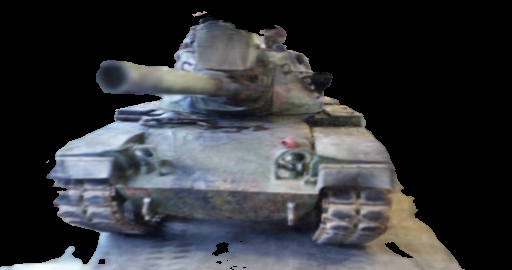}}
\end{minipage}

\begin{minipage}{0.2\linewidth}
\begin{tabular}{cc}
\multicolumn{2}{c}{\textbf{NERF}}                                                \\ \hline
PSNR & Time \\
21.62 & 27h35m
\end{tabular}
\end{minipage}
\begin{minipage}{0.8\linewidth}
\subfloat{\includegraphics[width=0.11\linewidth]{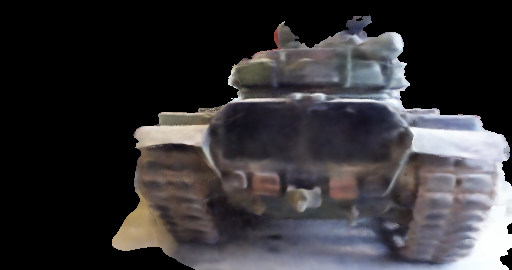}}
\subfloat{\includegraphics[width=0.11\linewidth]{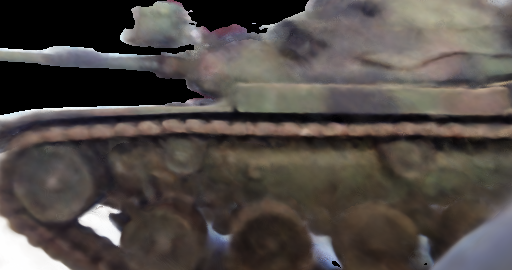}}
\subfloat{\includegraphics[width=0.11\linewidth]{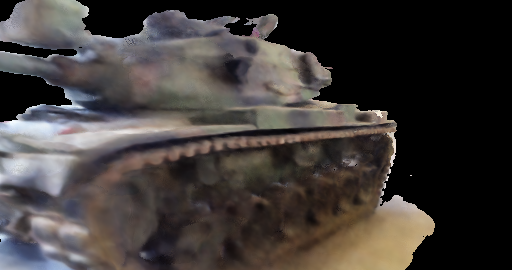}}
\subfloat{\includegraphics[width=0.11\linewidth]{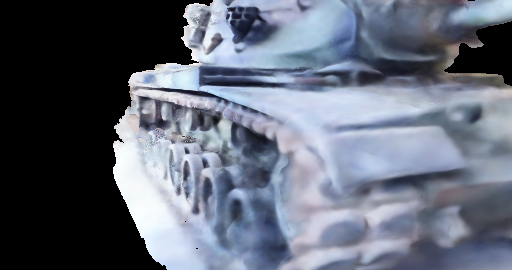}}
\subfloat{\includegraphics[width=0.11\linewidth]{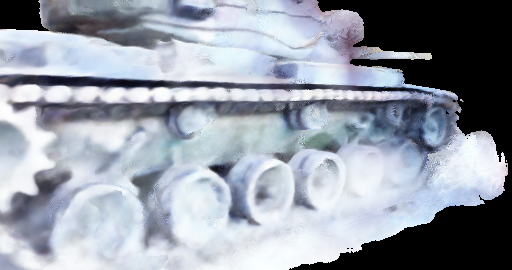}}
\subfloat{\includegraphics[width=0.11\linewidth]{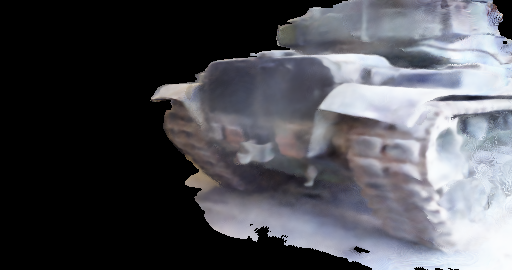}}
\subfloat{\includegraphics[width=0.11\linewidth]{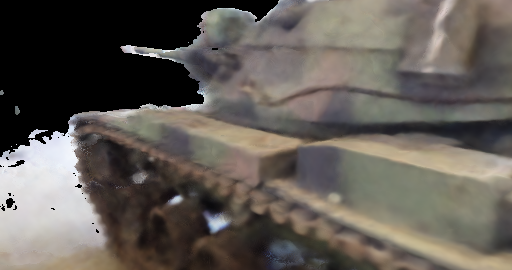}}
\subfloat{\includegraphics[width=0.11\linewidth]{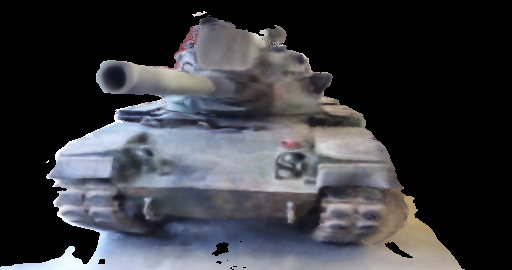}}
\end{minipage}

\caption{Results on the 360\textdegree scenes from the NERF data set as well scenes from the Tanks and Temples data set. Our reconstruction on eight hold-out views are shown compared to 
NERF without view-dependance. To the left final reconstruction times and PSNR-scores are shown for the corresponding scenes. The quality of our reconstructions are similar to those of 
NERF, while taking 2-3 order of magnitudes less time.}
\label{fig:real}
\end{figure*}


In the Figure~\ref{fig:real}, we have reconstructed the two 360\textdegree\ scenes used in the NERF paper, as well as a number
of scenes from the Tanks and Temples benchmark suite. To get a comparable metric, we have first masked out the foreground for 
each input image using our method. The PSNR metric have then been computed on the results of both our and NERF's 
reconstructions while applying these masks. We get similar quality using our method compared to the reference. By visual inspection, our result has a bit higher resolution but with more noise compared to NERF.

\subsection{Performance}
In the right side of Figure~\ref{fig:nerfsyn}, corresponding PSNR-curves are shown for our method versus 
the reference for the artificial scenes of the NERF data set.
We can see that the convergence is much higher for our method while achieving a similar or better final 
PSNR-score.

For the real scenes in Figure~\ref{fig:real}, the computation time is shown 
to the left for each scene for our method and the reference. Since proper comparison require the 
application of a foreground mask from our finished scene, we only show the final time until convergence 
for our method versus NERF. We can see that our method is 30 - 400 times faster than the reference 
method at comparable quality. The main reason for the spread in computation time for our method is if
it has required four or five hierarchical levels to achieve the same level of quality as the reference.
This in turn is largely decided by the relative distance between the cameras and the object of
reconstruction, since we use a fixed discretization using a voxel grid.

\section{Discussion}
We present a novel method that can be used to reconstruct both real and artificial 360\textdegree\ scenes.
We obtain comparable quality to the reference, both subjectively and with respect to PSNR (Figure~\ref{fig:nerfsyn} and Figure~\ref{fig:real}). Performance-wise, our reconstruction 
times are about two to three order of magnitudes faster then for the compared method (Figure~\ref{fig:nerfsyn} and Figure~\ref{fig:real}).
\\
In this paper, we chose to limit our reconstruction to view-independent radiance fields, as we focused on 
the explicit reconstruction of the non-linear least-squares formulation, and the task of solving this with
high performance.
Even though the reconstruction show
some robustness to view-dependant effects such as highlights, highly specular scenes from the NERF data 
set had to be omitted. This is true both for our method and for NERF without view-dependence. It would 
therefore be interesting to extend the ideas in this paper to incorporate view-dependance, to enable a
true comparison between our method and NERF.
\\ 
Another interesting idea to follow up on would be to exploit the fact that our method require about three
order of magnitudes fewer iterations until convergence (see Figure~\ref{fig:adam_vs_nerf}). This makes
our method a strong candidate for distributed computing, since that requires synchronization of
intermediate results between nodes for each iteration.


\ifCLASSOPTIONcompsoc
  \section*{Acknowledgments}
\else
  \section*{Acknowledgment}
\fi
This work was supported by the Swedish Research Council under
Grant 2014-4559.

\bibliographystyle{IEEEtran}
\bibliography{refs}

\end{document}